# Acoustic Scene Analysis using Analog Spiking Neural Network

Anand Kumar Mukhopadhyay, Naligala Moses Prabhakar, Divya Lakshmi Duggisetty, Indrajit Chakrabarti, and Mrigank Sharad

**Abstract**

Sensor nodes in a wireless sensor network (WSN) for security surveillance applications should preferably be small, energy-efficient, and inexpensive with in-sensor computational abilities. An appropriate data processing scheme in the sensor node reduces the power dissipation of the transceiver through the compression of information to be communicated. This study attempted a simulation-based analysis of human footstep sound classification in natural surroundings using simple time-domain features. The spiking neural network (SNN), a computationally low-weight classifier derived from an artificial neural network (ANN), was used to classify acoustic sounds. The SNN and required feature extraction schemes are amenable to low-power subthreshold analog implementation. The results show that all analog implementations of the proposed SNN scheme achieve significant power savings over the digital implementation of the same computing scheme and other conventional digital architectures using frequency-domain feature extraction and ANN-based classification. The algorithm is tolerant of the impact of process variations, which are inevitable in analog design, owing to the approximate nature of the data processing involved in such applications. Although SNN provides low-power operation at the algorithm level itself, ANN to SNN conversion leads to an unavoidable loss of classification accuracy of ~5%. We exploited the low-power operation of the analog processing SNN module by applying redundancy and majority voting, which improved the classification accuracy, taking it close to the ANN model.

## 1. Introduction

Wireless sensor networks (WSNs) have applications in a wide range of health care, environment, agriculture, public safety, military, industry, and transportation systems [1]. The sensor nodes in a WSN are designed according to the target application. Acoustic sensor nodes for a security surveillance application to monitor the presence of human beings in restricted zones fall under the category of passive supervision, which senses the data around its environment and requires minimal power to operate. In general, small-sized static sensor nodes are deployed in large numbers with continuous monitoring. These resource-constrained sensor nodes should be inexpensive and power-efficient (few $\mu$W) for serving the purpose [2].

One of the constraints that the wireless sensor nodes suffer from is limited battery life. The most power-hungry component in a WSN is the transceiver unit [2, 3]. Hence, in-sensor computing is applied in the sensor nodes to minimize the communication power due to compressed data transmission. The sensor nodes having in-sensor computational abilities should consume less energy, preferably much lower than the transceiver unit. Therefore, there is a need for a cross-hierarchy power-efficient system design approach that spans the application-specific computing systems algorithm, architecture, and circuit-level implementation.

With the advancement in the field of energy-efficient brain-inspired neuromorphic computing algorithms, several low-power computing systems have been envisaged and implemented, which are suitable for in-sensor computing and data classification [4][5][6][7][8][16][55][60]. Spiking neural networks (SNNs), considered the 3rd generation of neural networks [38], are inspired by biological neurons in which communication between neurons takes place in the form of spikes (or events) through interconnecting 'weights' called synapses. The neuron membrane potential ($v_{mem}$) is affected based on its presynaptic events, i.e., because of the spikes generated by the preceding layer neurons. A post-synaptic spike is generated whenever $v_{mem}$ exceeds the neuron's threshold potential ($v_{th}$), and communicated to the next layer of neurons connected to it. Such bio-mimicking neural networks significantly reduce computational complexity, owing to event-driven computing and binary signal levels[39].

The learning rules for SNN algorithms can be supervised or unsupervised, rate-based, or spike-time based, which is chosen based on the application [9]. The primary motivation for involving SNN in a real-time application involving energy-efficient neuromorphic hardware is to reduce the computations involved in the system [55][62]. Generally, the computations during the training of neural network frameworks are compute-intensive compared to the feedforward inference process. Hence, offline training with online inference is suitable for a static sensor node and its associated classifier [40][57]. We have considered a method for training an ANN with the conventional backpropagation algorithm to obtain the trained parameters using a rate-based SNN for classification [6]. This approach achieves acceptable classification performance owing to offline supervised training with a more extensive data set and low-power operation due to lesser computations involved in the SNN-based classifier.

In this work, we present an integrated system design for a low-power sensor node targeted for a surveillance application. The node is supposed to spot human presence in restricted areas by analyzing the generated acoustic audio signals using simple time-domain features, analog-domain processing, and SNN-based neuromorphic classification [11][12]. We compare the algorithm level performance of the proposed scheme with standard techniques and provide architecture and circuit level design for low-power implementation of the proposed system and compare it with the conventional methods. The strategy for improving the classification accuracy for the SNN classifier is also explored, which makes it comparable to the traditional ANN classifier, in terms of performance. The significant contributions of the work are:

- To design and optimize time-domain features extracted from acoustic data for SNN-based processing.
- To design and optimize SNN parameters for improved performance.
- To compare the performance of the proposed method with the conventional method consisting of a frequency-domain feature extractor and ANN classifier.
- To propose and implement a low-power mixed architecture scheme and compare it with the digital implementation of the same scheme and conventional approaches.
- To assess the variation tolerance of the proposed mixed neuromorphic implantation.
- To improve the SNN inference module's performance by redundancy scheme and majority voting.

The rest of the paper is organized as follows: In **section 2**, a brief description of the application is given, along with a review of some conventional approaches addressing similar applications. **Section 3** presents the methods to explain the proposed scheme based on time-domain feature extraction and SNN-based classification, algorithm level optimization of the feature extraction scheme and the SNN-based classifier, and the proposed scheme's digital and mixed architectural implementation. In **section 4**, simulation results for the proposed solution and its comparison with conventional approaches are discussed. **Section 5** concludes the paper by highlighting the key takeaways and scope for future work.

## 2. Application overview and related works

This section presents the overview of the application explored in this work, namely, a low-power acoustic sensor node, with in-sensor computing, for surveillance. The objective is to detect the presence of humans in a restricted zone. This smart sensor system has importance in military surveillance applications for monitoring human existence in sensitive areas. A block-level diagram of the neural network-based sensor node is shown in **figure 1**, which consists of units for power management, sensor interface, filtering, analog to digital conversion (ADC), data processing, memory, and wireless communication. The sensor node is supposed to capture the surrounding sound signals within its range, which would be processed on the device using the proposed low-power computing scheme and communicate the results to a base station.

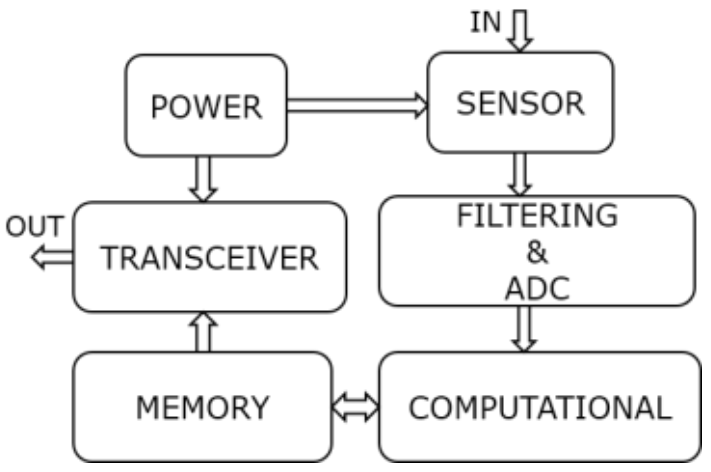

**Figure 1**. The major blocks required in a wireless sensor node system.

Conventional audio signal feature extraction methods usually rely on a spectral, wavelet, and Mel-frequency cepstral coefficients (MFCC) features from the frequency domain [41, 42], mean standard deviation, and the maximum amplitude value from the time domain. These features are the input vectors to a machine learning or neural network model for classification [33-37].

A detailed review of acoustic signal processing and the different stages involved (sound rejection, detection, classification, and cancellation) is presented in [33]. In [34], FFT features from the acoustic and seismic signals were considered along with baseline classifiers such as KNN (K-nearest neighbor), maximum likelihood (ML), and support vector machine (SVM) for vehicle classification. In [36], spectral features use linear prediction coding (LPC) and a neural network to classify seismic signals. In [37], both time and frequency domain features were considered and assessed using neural networks and genetic algorithms for classifying seismic signals. In [12], a time-domain signal processing for feature extraction, followed by a neural networks-based classification of acoustic/seismic signals resulting from different types of vehicles

, is studied. A review of sound event recognition for audio surveillance describes the importance of sound classification tasks in noisy environments using machine learning and neural network methods [42]. Indoor security monitoring for classifying audio using cepstral-domain-based features and time delay neural network is studied in [44]. A time-delay neural network (TDNN) was used to classify footsteps, fan noise, wind noise, door opening, and door close using MFCC for indoor security monitoring [44]. The TDNN classifier and MFCC feature detector have high complexity and will consume more resources and power. MFCC, energy, pitch range (PR), and linear prediction coefficients (LPC) feature, along with standard ML models such as KNN, support vector machine (SVM), and Gaussian Mixture Model (GMM), are considered for forensic audio analysis and recognition of criminal suspects using footstep sound signals in [48] [49]. MFCC extracts valuable information from the acoustic signals trained using standard ML models for better results. However, for security surveillance in remote locations, low complexity classifier and feature extractor must be designed due to the limited battery life of the in-sensor systems.

Time-domain, spectral, and cepstral features derived from multiple microphones using support vector machines (SVM) classifier are studied for indoor footstep detection [46] [47]. The multi-channel approach improves the noise robustness. In [46], authors have considered six-time domain features such as peak amplitude, peak power, time to the next event, attack time, decay time, and zero-cross rate; 7 spectral features containing information on the amplitude and frequencies for 1st -3rd most significant the number of peaks; 24 MFCC features. Such systems are robust because it considers extensive feature vectors on the input data captured at different angles from multiple microphones; hence, all possible predictors should be used for superior performance for systems with unlimited power access. Both [44] and [46] apply to indoor security surveillance systems.

However, it is essential to identify valuable features and classifiers with low complexity for power-constrained portable systems. In [58], authors have shown how SNNs can provide high performance at low energy cost for human activity recognition problems which could be extended to multiple domains involving on-edge artificial intelligence applications. This work aims to design an approximate innovative sensor node system suitable for extracting essential information from the time domain and a low-weight bioinspired analog SNN classifier [5, 8, 12, 30, 45].

This work aims to develop algorithms and the corresponding circuits and architecture for low-power edge computing on miniature, power-constrained sensor nodes. The aforementioned conventional methods suffer from high computational complexity in feature extraction and classification, leading to relatively power-hungry solutions. Therefore, customized, application-specific, energy-efficient techniques for real-time feature extraction and classification would be necessary, leveraging the approximate nature of the processing required for the application of interest. For instance, in this work, we target a classification scheme for making a binary decision: the absence or presence of humans in an acoustic scene through sounds produced due to motion. The acoustic signals produced by human movement may be limited to specific frequency bands that would be quite distinct from other natural noise sources. We may be able to deploy relatively less complex feature extraction methods and classification schemes rather than adopting compute-intensive frequency transforms [11, 13]. We approached the problem considering low-weight time-domain feature extraction and bio-inspired SNN-based classification. This is an approximate computing methodology amenable to low-power, analog implementation.

## 3. Methods

### *3.1 Time Domain Approximate Feature Extraction and SNN Based Classification Scheme*

We propose a scheme involving low complexity time-domain features followed by an SNN classifier as a promising alternative for low-power sensor node design for acoustic applications. The time-domain features reduce computational complexity compared to conventional frequency domain analysis [42]. The feature set used in this work is based on signal length, zero-crossing, and local signal extrema count. It inherently captures an approximate statistic of high and low-frequency components in the target signal. The low-frequency content is expressed by counts of signal samples within consecutive zero crossings, while the high-frequency range is reflected by the counts of signal extrema on either side. In addition to that, the SNN classifier is employed as an alternative to its counterpart ANN due to the spike (event) based processing of information, aiding low-power hardware realization.

### *3.1.1 Low complexity time-domain feature extraction*

Features derived directly from the time domain would be preferable in terms of lesser computation complexity when compared to frequency domain features. Authors in [12] have earlier shown that time encoded signal processing and recognition (TESPAR) has successfully obtained essential elements for seismic signals and hence is feasible for real-time processing.

TESPAR is inspired by infinite clipping coding of a waveform which represents the duration between the zero crossings of the waveform [12]. The segment between two consecutive real zeroes is termed an epoch. The intelligibility of the waveform is enhanced by introducing the concept of complex zeroes in addition to the real zeroes. The complex zeroes represent the shape of the signal waveform in the form of local maxima and minima of the signal waveform. Therefore, a bandlimited signal is approximated by computing the following parameters in each epoch, as shown in **figure** 2(a).

The D and S contain information about the signal's fundamental frequency and harmonics [12]. A two-dimensional (2D) D-S matrix representing a signal waveform is shown in **figure** 2 (b). Each location in the 2D matrix will have the number of occurrences of the particular D-S combination present in the signal during the external time window (ETW). An ETW of 5s is sufficient to qualify the incoming signal as a data sample (discussed in **section 3.2.1**). For instance, the value entered in D-S (2,1) would be the count of the D-S pair with value (2,1) in the last ETW. The 2D matrix is further expanded to a 1D vector to form the classifier's final feature vector (FV). The optimization of different parameters related to it is discussed in **section 3.2**. Note that the authors in [10] have applied a feature vector generation scheme, which involved several D-S pairs having counts in different ranges. For instance, a number of D-S pairs having count '0' are entered as the 1st element of the feature vectors, the 2nd element would be the number of D-S pairs having count '1', and so on [10]. This would involve significant overhead in memory read-write operations. Hence, we have employed the D-S pair counts directly as feature vectors for SNN-based classification in this work.

Moreover, we have shown that application-specific optimization further reduces the complexity by selecting only the significant D-S pairs for the feature vector. Several other algorithmic parameter optimizations are discussed in **section 3.2.**

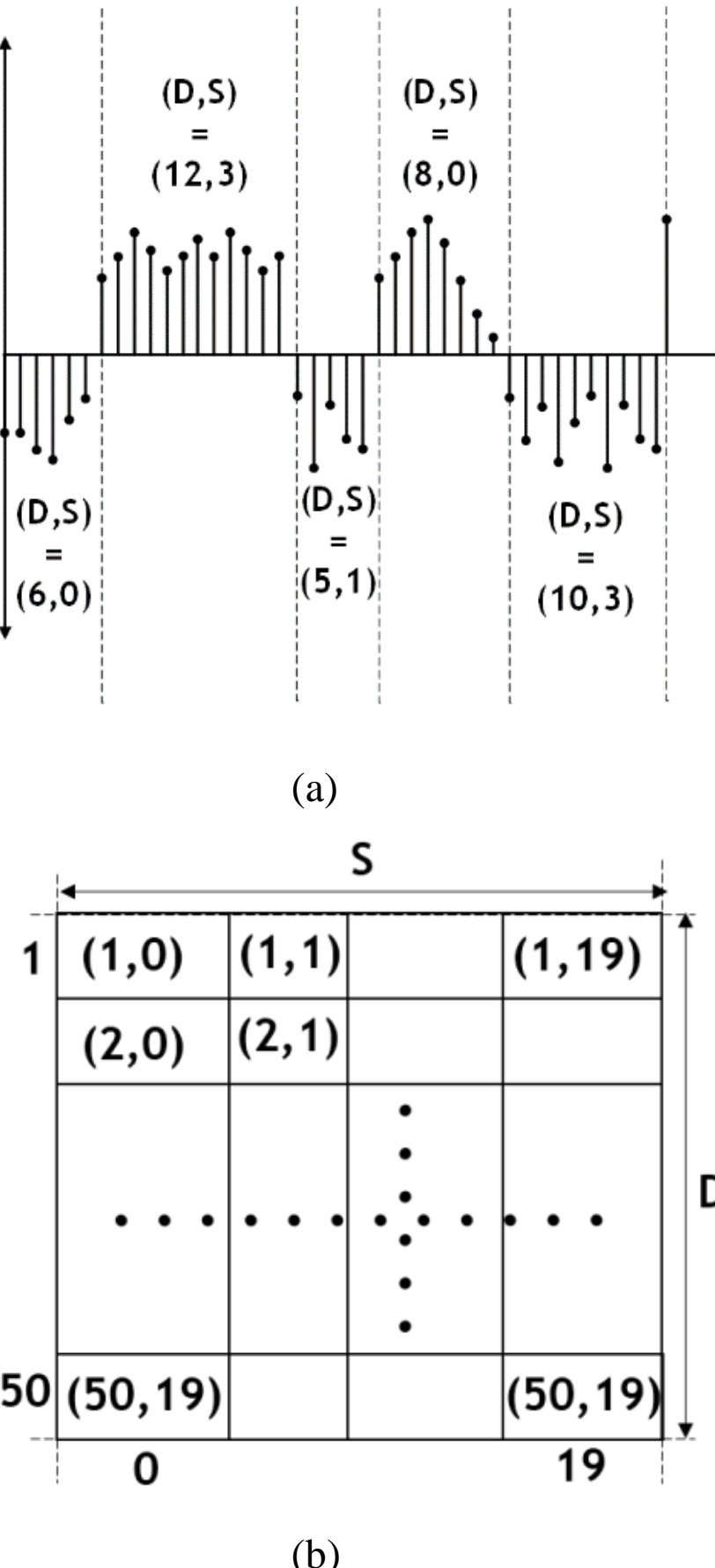

**Figure 2**. (a) Small portion of the sampled audio signal, (b) 2-D matrix representation of D and S [Duration (D): It is the number of samples between real zeroes (within an epoch, Shape (S): It is the number of local minima and local maxima for a positive and negative epoch, respectively (S=0 if no maxima/minima present)].

Low-power, analog implementation of the proposed scheme is presented in **section 3.4.2**.

### *3.1.2 SNN based classification scheme*

The final feature vector is fed to an SNN classifier. The general network structure of an SNN is like that of the ANN, as shown in **figure 3(a)**. The difference lies in the functionality of the neurons present in the network.

An illustration of the spike generation process from a feature vector within a specified time window is shown in **figure 3(b)**. The number of spikes generated in the interval is directly proportional to the FV magnitude and the maximum input firing rate ($r$), a network parameter. The input layer neurons are processing spikes equivalent to the FV value.

Integrate and fire (I&F) neurons are used in the subsequent layers of the SNN to implement the spiking operation, as shown in **figure 3(c)**. The I&F neuron receives input from the previous layer neurons whenever there is a presynaptic spike event. The neurons between two-consecutive layers are connected via synapses of different strengths representing the weights. Hence, the input current magnitude is scaled down according to the synaptic strength. After receiving the input, the neuron's membrane potential ($v_{mem}$) increases. Whenever $v_{mem}$ exceeds the threshold potential ($v_{th}$) of the neuron (i.e., $v_{mem}>v_{th}$), the neuron is activated. It generates a post-synaptic spike, after which $v_{mem}$ falls to the neuron resting potential ($v_{res}$). The mathematics depicting the $v_{mem}$ of a simple I&F neuron is shown in **equation (1)**.

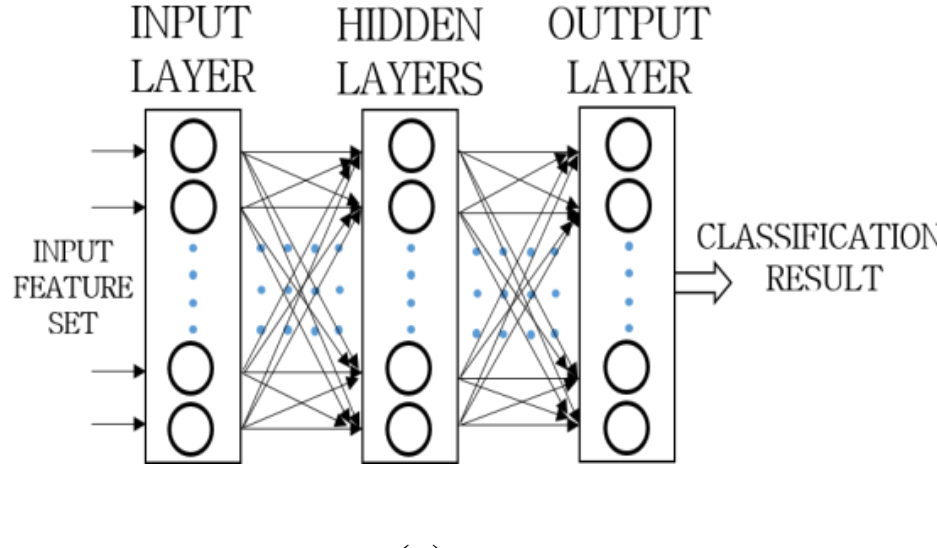

(a)

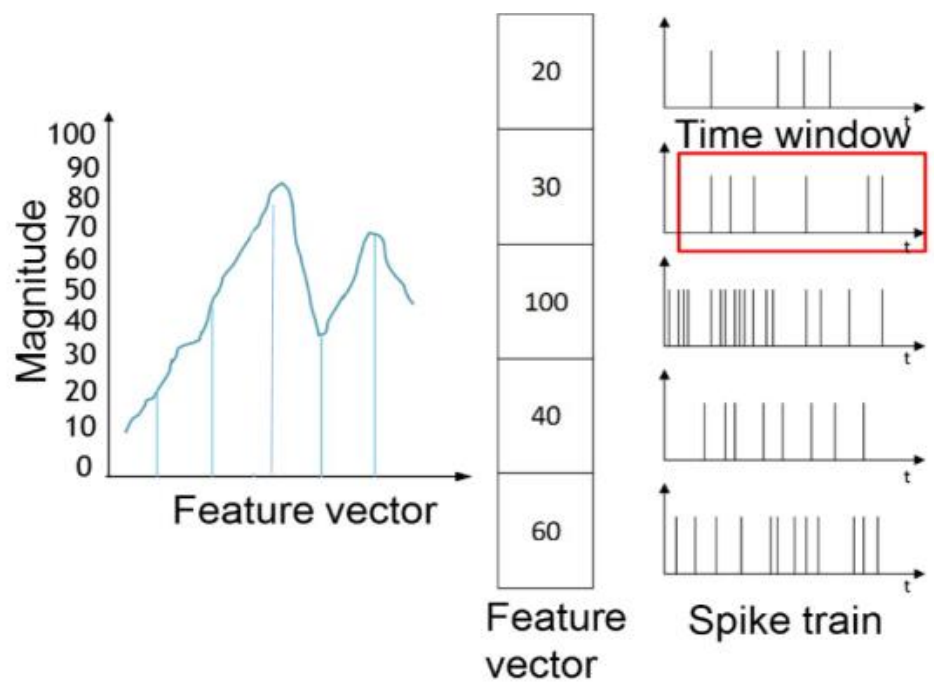

(b)

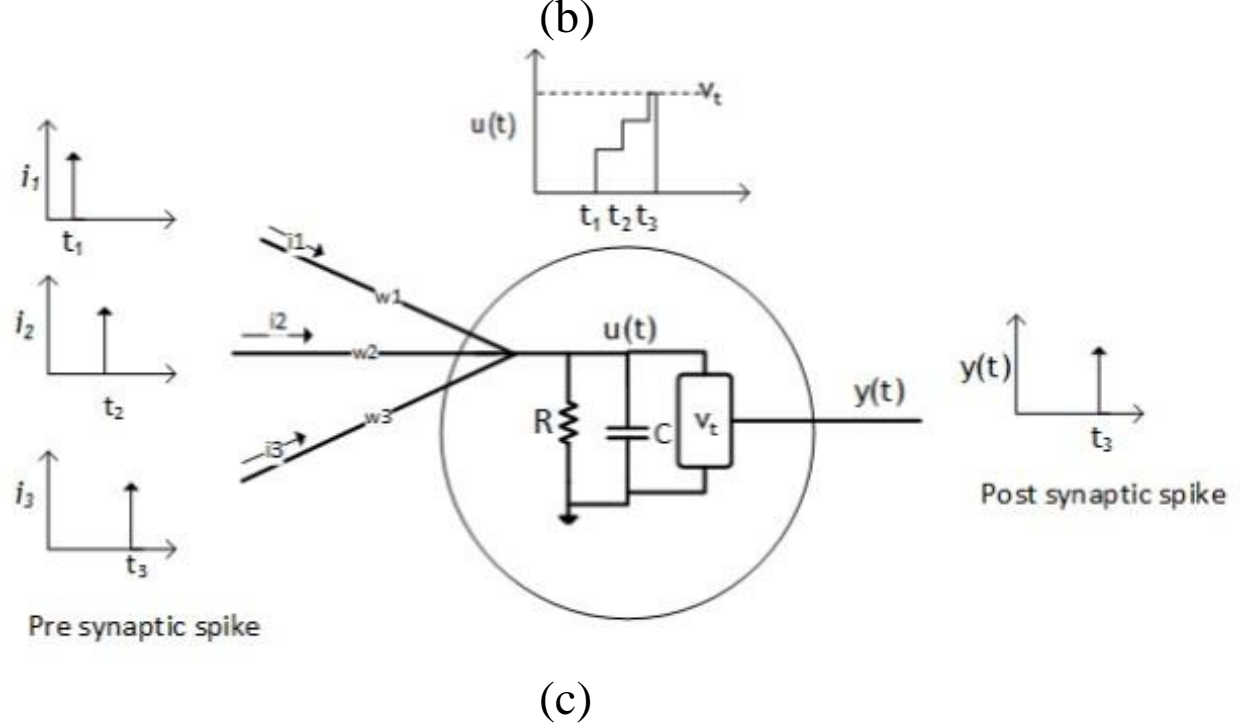

(c)

**Figure 3.** (a) Network structure of the neural network (ANN/SNN) model, (b) Illustration of the spike generation process, (c) Schematic diagram of an integrate and fire (IF) neuron model

$$\frac{dv_{mem}(t)}{dt} = \sum_{i} \sum_{a \in S_i} W_i . \delta(t-a) + v_{res} \qquad (1)$$

Here, $W_i$ and $\delta(.)$ represent the weight of the $i^{th}$ incoming synapse and delta function, respectively. Si = [$t_i^0$, $t_i^1$, …..] stores the $i^{th}$ presynaptic neuron spike times.

The optimal choice of network parameters for the ANN and the corresponding SNN network is discussed in **section 3.3**. The architectural diagram of the SNN classifier is constructed using digital and analog fashion (refer to **section 3.4**). The model in equation (1) is particularly suitable for low complexity analog implementation, with the help of physical characteristics of devices like transistors, capacitors, and simple sub-threshold circuits [43]. Hence, in this work, we propose an implementation for the SNN classifier, which interfaces well with the analog mode feature extraction module.

### *3.2 Optimization of feature extraction parameters*

Optimization of features is necessary to achieve an acceptable classification accuracy with minimal computational operations in the feature extraction and the subsequent classifier module present on the sensor node. The following subsections describe the feature engineering process on the acoustic signals used in this work.

#### *3.2.1 Choice of Running Window ($R_{WIN}$) and External Time Window (ETW)*

**Figure 4** illustrates the running and external time window (ETW) of 5s duration. The external time window (ETW) determines the signal length to qualify it as a data sample, and the running window size ($R_{WIN}$) is the time step for shifting or sliding the ETW. The length of $R_{WIN}$ determines the throughput of the system. If $R_{WIN}$ is too short, the changes in the feature vector will be insignificant, leading to redundant results. On the contrary, too large $R_{WIN}$ would lead to insufficient overlap between the two consecutive time segments, which would, in turn, increase the possibility of missing features of interest spanning over two such elements. A smaller ETW will fail to capture a real-time footstep signal occurring in the vicinity. In contrast, a large ETW would include substantial background noise and diminish the effect of a small duration footstep signal. Hence, choosing the best ETW size is highly correlated with the average duration of the footstep signals, and too large or too small ETW adversely affects the classification accuracy. $R_{WIN}$ = 500ms and ETW = 5s were decided based on classification accuracy as studied in [59].

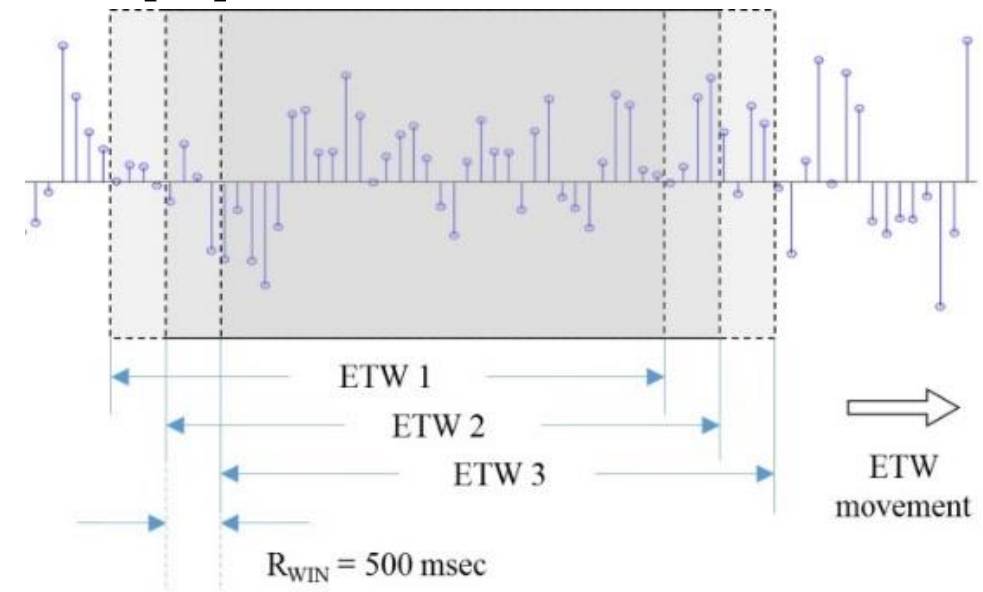

**Figure 4.** Illustration of the Extended Time Window sliding across the sound waveform.

### 3.2.2 Dataset creation

The data samples for our analysis were created considering the sensor nodes in a gravel surface with an ETW=5s. The human footstep audio clips were mixed with different background noises like those resulting from heavy rain, chirping birds, blowing of the wind, and crickets stridulating. We created a suitable dataset[1] using data augmentation methods such as rolling the files circularly and adding random noise due to the unavailability of a significant number of data needed to address the problem [14]. The footstep sound on gravel and background noise audio data were taken from online sources sampled at 44.1kHz. From the frequency spectrum of the audio files, it was understood that a significant portion of the power content lies in the frequency range of (0-5) kHz, both for the background noise and footstep sound. Therefore, a cut-off frequency of $f_c$=5kHz was chosen for filtering out the high-frequency signals, and hence we downsampled the data to a sampling frequency of 11kHz. The magnitude of power content depends on background noise or footstep intensity, which may vary in different instances. Too high background noise would make the detection difficult, eventually leading to failure. The range of footstep and background power considered are (39.85-44.61) dB and (43.73-45.14) dB, respectively. Hence, the dataset consisted of signals with $(SNR)_{dB}$ as low as -5.29dB, an acceptable value for our application [44]. It is to be noted that appropriate filters must be used to reject the unwanted noise for lower $(SNR)_{dB}$ due to high noise power in the actual scenario.

The task is to detect human presence on a fixed surface (gravel considered) in the presence of natural environmental sounds; four different background noises were considered, namely, 1) crickets, 2) birds chirping, 3) rain, and 4) wind. For the human presence, audio files of a single person walking on gravel were acquired. The file was divided into 5s segments, considering it a suitable ETW for real-time analysis. For emulating multiple people (>3) walking on gravel, signals of a single person walking were combined with a random scaling factor between 0 to 1 to incorporate the effect of distance or intensity from the sensor node location. The final dataset created consisted of 12352 data samples consisting of two classes, 1) background noises (6176 samples), 2) single person and multiple people walking on gravel in the presence of background noises (6176 samples). The total dataset was divided into a train set (9024 samples), validation set (1664 samples), and test set (1664 samples) for further analysis. The number of a single person and multiple people, including variation in walking speed and the four different background noises, were selected in a balanced manner for incorporating their effects in equal proportion during the network training.

The footstep audio signal is initially filtered with a low pass filter with a passband frequency 5KHz, and stopband frequency 6KHz as the footstep audio signal lies within the frequency range (0-5) kHz. Then, different sampling frequencies ($f_s$) were chosen, and their effect on classification accuracy was observed while setting the (D, S) parameters to (20, 5). It is to be noted that the test data samples for different frequencies are duplicate files but resampled to different frequencies. It is observed that the change in test accuracy is insignificant even at lower $f_s$. Hence, $f_s$ = 11025Hz has been chosen to reduce the computational complexity for further analysis.

### 3.2.3 Choice of D and S matrix dimension

The two-dimensional matrix of the (D, S) combination for (50, 20) and (20, 5), as shown in **Figure 5**, indicates that most of the essential features are concentrated in lower values of (D, S) combinations. The test accuracy observed for different (D, S) combinations at $f_s$ = 11025Hz is shown in **Table. 1**. It is evident that an acceptable accuracy is obtained even at a lower $(D_{max}, S_{max})$ combination. As mentioned earlier, the feature vector (FV) used for classification is simply the elements of the D-S matrix. The FV length equals D × S, which determines the number of neurons in the neural network's input layer. A lesser number of input neurons is preferable due to the reduction in the architectural size of the classifier. Therefore, a (D, S) combination of (10, 5), implying a feature vector of dimension 50, is considered, which provides an accuracy of ≈ 93.32%. Further reduction in the size of the D-S matrix led to a significant decrease in the classification accuracy.

### 3.2.4 Bit-length of Feature Vector (FV)

The range of values of the FV elements is shown in **figure 6 (a) (b)**. Some features have a very high value that nullifies the effect of the other components or would require

**Table 1:** Effect on test accuracy for different ($D_{max}$, $S_{max}$) Combinations at $F_s$ = 11025Hz.

| (Dmax, Smax) | FV dimension (number of input layer neurons) | Test accuracy (%) |
|---|---|---|
| (5,5) | 25 | 92.30 |
| (10,5) | 50 | 93.32 |
| (20,5) | 100 | 93.44 |
| (30,5) | 150 | 93.32 |
| (50,5) | 250 | 91.40 |

[1] Dataset will be provided upon request.

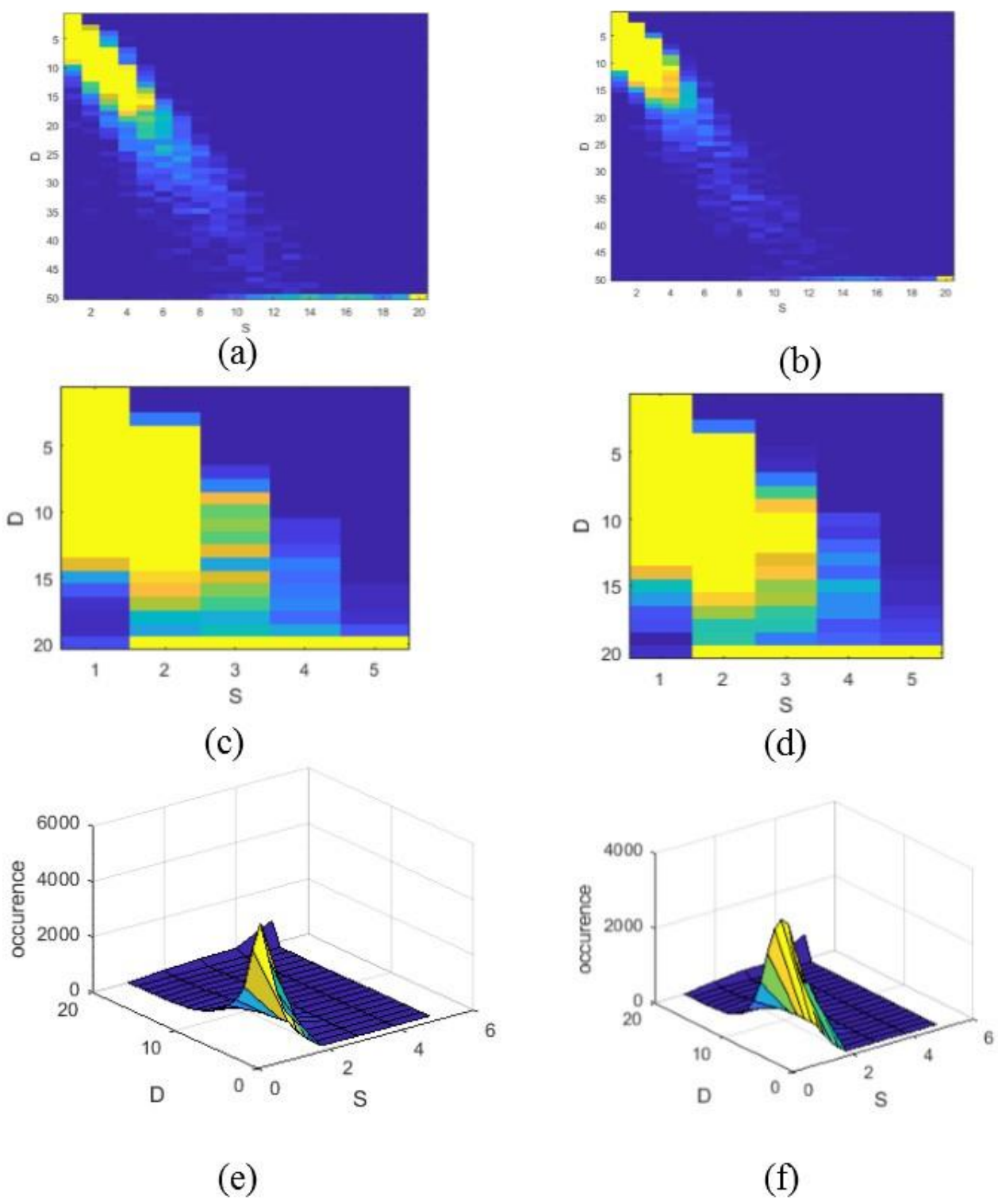

**Figure 5.** DS matrix 2-D representations (a) chirp of birds (D=50, S=20), (b) single person walking on gravel with a chirp as background noise (D=50, S=20), (c) chirp of birds (D=20, S=5), (d)single person walking on gravel with a chirp as background noise (D=20, S=5), (e) surface view of (c), (f) surface view of (d)

a large number of bits to accommodate it. A more significant number of bits would imply higher computational complexity. One possible solution would be to employ a log scale. Another more straightforward approach is to limit the count for the high-frequency FV elements to a maximum value, provided it does not deteriorate the performance. Simulation-based analysis showed a steep fall in the classification accuracy if the maximum value was reduced below 256 (8 bits), as seen in **figure 6 (c),** as the high-value features approach is closer to the lower magnitude features and reduces the signal's distinctive characteristics. The effect on test accuracy due to the FV bits is shown in **figure 6 (d)**. Henceforth, we represent the FV elements using 8 bits for further analysis.

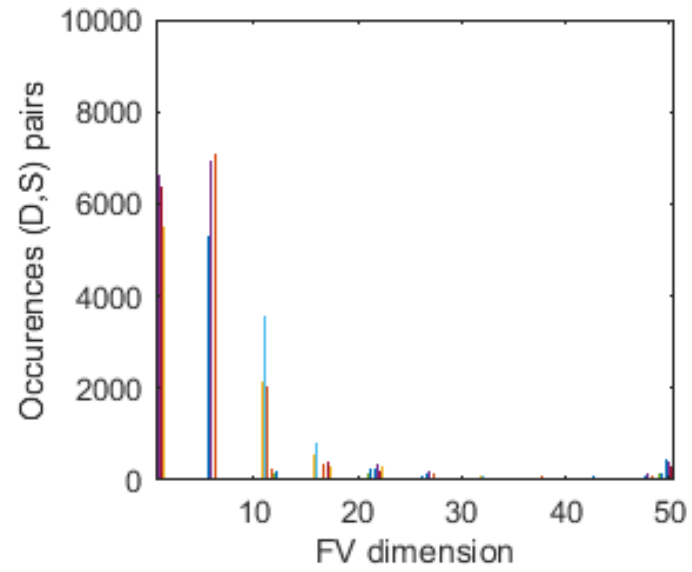

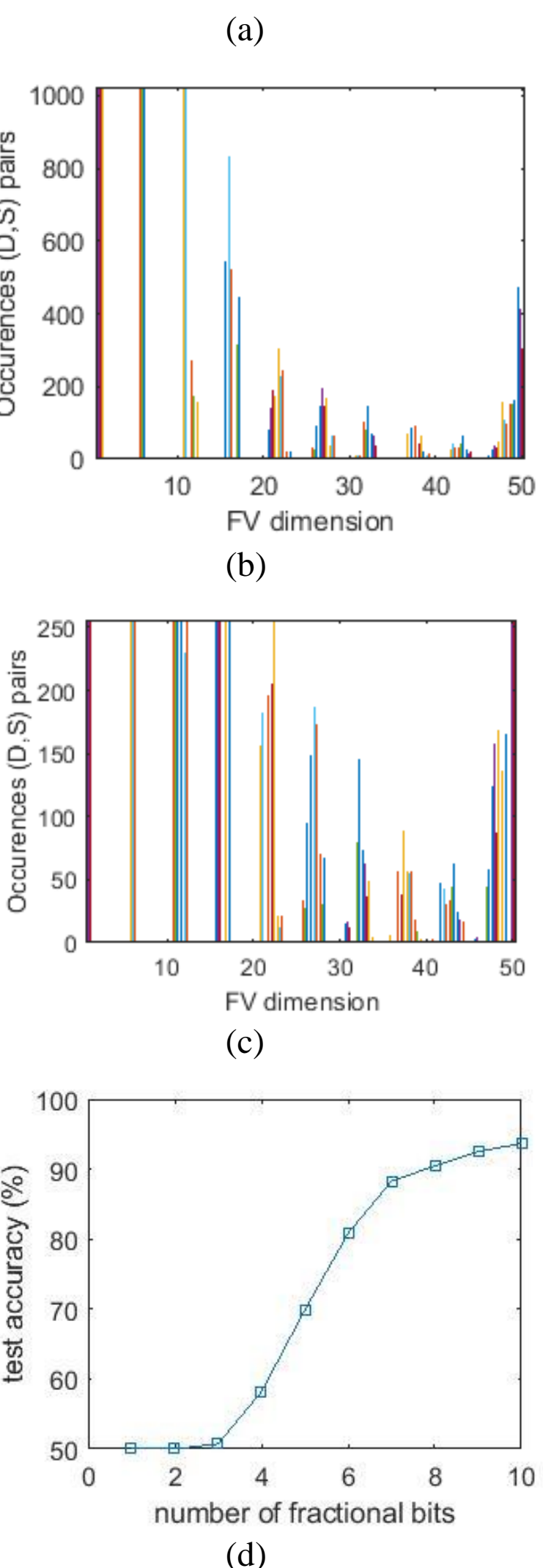

**Figure 6.** (a) The FV (50x1) magnitudes for ten different samples (a) original range (b) maximum value set to 1024, (c) maximum value restricted to 256, (d) effect on accuracy versus the number of bits used in the FV.

### *3.3 Optimal Conversion of ANN to SNN*

In **section 3.1.2**, we have explained the SNN-based classification scheme. It is understood that the FV is represented by binary spike trains with random '1's and '0's, and the number of spikes (i.e., '1's) is proportional to the magnitude of the FV element value. Whereas in the case of ANN, the FV values are real-valued. A random number is generated at every step of the SNN operation, compared to the magnitude of the corresponding input FV. A spike event is triggered if the generated random number is less than the input FV element. This process makes sure that the magnitude of FV presented as ANN inputs are proportional to the number of input spikes for the corresponding SNN [5, 6]. The spike-based operation has two main advantages: 1) It eliminates the multiplication operation between the FV and weights as required in ANN. The synaptic weights are simply added and accumulated whenever a spike is generated in the presynaptic neuron. This accumulated value is the membrane potential ($V_m$) of the post-synaptic neuron, which produces a post-synaptic spike if, $V_m > V_t$, $V_t$ being the neuron firing threshold potential, and 2) The addition of the synaptic weight between the pre and post-synaptic neurons is skipped if the previous layer neuron does not produce a spike. In addition to these algorithm-level benefits, SNN is also suitable for low-power sub-threshold analog implementation.

The mathematical model for the output of neurons in a deep neural network (DNN) is given by **equation (2)**

$$y_i^{(l)} = F\left(\sum_{n=1}^{N} W_{ij}^{(l)}.(FV)^{(l-1)} + b_i^{(l)}\right) \quad (2)$$

Here $y_i^{(l)}$ is the output of neuron $i$ in layer $l$, where $l \in [1....L]$ denotes the $l^{th}$ layer. $W_{ij}^{(l)}$, $b_i^{(l)}$, and $(FV)^{(l-1)}$ are the weight parameters, bias parameters, and the input feature vectors which connect neuron $i$ in the current layer to the neuron j in the previous layer, respectively. $F(.)$ is the activation function such as sigmoid, tanh, Rectified Linear Unit (ReLU). Here, ReLU is considered because it helps in faster convergence, is computationally efficient, and is required to map ANN to SNN [5, 35].

For the ANN to SNN conversion, the network is initially trained using a backpropagation training algorithm to minimize the cross-entropy loss function, as shown in **figure 7(a)** and given in **equation (3)**. The corresponding training and validation accuracy is shown in **figure 7(b)**.

After choosing the optimum FV parameters (**section 3.2**), viz., $f_s$ = *11025Hz, D = 10 and S = 5* we train an ANN with different network configurations. The following parameters were chosen during the training of the ANN model, learning rate, α = 0.001 with a categorical cross-entropy loss function suitable for the classification problem, Adam optimizer, a dropout (d = 0.5) is considered for avoiding overfitting. The backpropagation algorithm was used in training to minimize the cross-entropy cost function as shown in equation (3).

$$E = -\sum_{n=1}^{N} \hat{y}_n \log(y_n^{(L)}) \quad (3)$$

The variation in test accuracy for the trained ANN model by varying the network parameters, viz., 1) the number of hidden layers, 2) the number of neurons in the hidden layer, is shown in **figure**s **7(c) and (d)**. Henceforth, we choose the optimum network configuration, i.e., ([50(input)-80(hidden) - 2(output)] neurons for further analysis.

Here, $\hat{y_n} \in \{0,1\}^n$ is the labels (ground truth) associated with the training data in one hot coding format, and $y^{(L)}$ is the output of the model.

For the proper conversion of ANN to its corresponding SNN network, certain conditions, namely, 1) setting the bias to zero and 2) using ReLU activation functions, were considered [6]. ReLU provides a better approximation to relate the firing rate of the integrate and fire (IF) neurons in the SNN model. In contrast, a zero bias will eliminate the effect of external influence on the IF neurons, i.e., only the weights and the neuron threshold potential ($V_t$) will matter.

The SNN inference model uses the weights obtained after training the ANN model using backpropagation after conversion. The SNN inference model is optimized further with the SNN parameters: input firing rate *(r)* and neuron threshold potential *($V_t$)*. A higher $V_t$ implies that a neuron is less likely to spike as the neuron membrane potential will need more time to reach $V_t$. A higher $r$ is directly proportional to the input firing rate according to equation (4).

$$I_{spike} = \begin{cases} 1 & ,FV_i \geq sf \times rand() \\ 0 & ,FV_i \leq sf \times rand() \end{cases} \quad (4)$$

Here, the scaling factor (sf) is equal to $sf = 1/(dt \times r)$. A smaller simulation time step ( $dt$ ) will increase the sf and reduce the number of input spikes per timestep. $FV$ and $rand()$ are the input feature vector and random number generator normalized between 0 and 1, respectively. Hence, a proper combination of $r$ and $V_t$ will ensure higher accuracy, as shown in **figure 8(a)**. It is observed that $r = 1000Hz$ and $V_t = 1mV$ with $dt = 1ms$ attains an accuracy of 86.75% within 0.1s time duration ( $T_{dur}$ ), as shown in **figure 8(a)**. The number of spikes generated in the hidden layer neurons ($N_{spk_hl}$) based on the different ($r$, $V_t$) pairs, is shown in **figure 8(b)**. At firing rates higher than 1000Hz, it is seen that $N_{spk_hl}$ reduces for a particular $V_t$. At lower $V_t$, the $N_{spk_hl}$ increases, but the accuracy falls. This is because the number of spikes in the hidden layer increases significantly at lower neuron thresholds, which will misinterpret the input FV appropriately. Hence, a proper combination of ($r$, $V_t$) and other design parameters ($dt$, $T_{dur}$) provides the best solution.

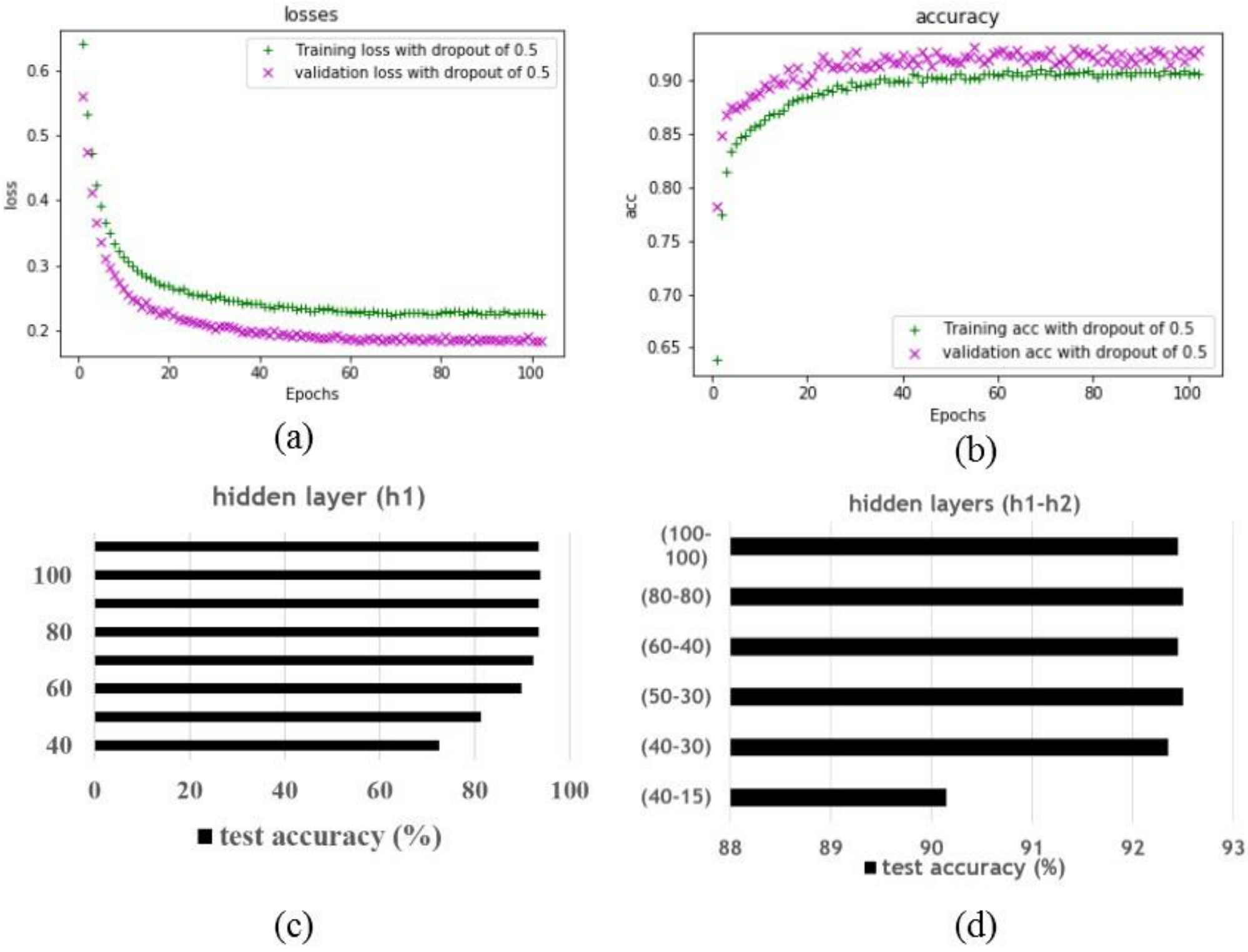

**Figure 7.** (a) Loss function with epochs, (b) accuracy versus epochs, (c) Variation in test accuracy for the trained ANN model by varying the number of neurons in hidden layer 1, (d) hidden layer 1 and 2.

The accuracy plot of the SNN network with time for the best case is shown in **figure 8(c)**. The average number of spike occurrences for the IF neurons in the input, hidden, and output layers are shown in **figures 8(d), (e), and (f)**, respectively. From the statistics of spike generation, the number of spikes generated in the input, hidden, and output layers are 343, 330, and 31, which sums up to ~ 700 spikes for the SNN operation for 100 iterations ($N_{iter}$). It is to be noted that the neurons that do not produce a spike during the classification process are in an OFF state and help reduce power consumption, which is not possible using an ANN classifier.

For a network architecture of [50-80-2], sequential operation of neurons will need ~ [(50x80) + (80x2) = 4160] MAC operations for computations. Considering the worst case (assuming all the neurons are spiking at every iteration) for SNN, we require ~ [{(50x80) + (80x2)} x $N_{iter}$] addition operations. Considering the above statistics of average spike generation for $N_{iter} = 100$ (i.e., ~6.86% (343/5000) for input layer neurons and ~4.125% (330/8000) for hidden layer neurons), the number of addition operations ~ [(343x80 + 330x2) = 28100] for the duration of 100 iterations.

The floating-point weight coefficients are converted to fixed-point representation to reduce the memory footprint due to the storage of trained parameters. From **figure 9(a)**, the test accuracy falls from 93.26% to 60.87% when the number of fractional bits is reduced from 2 to 1. Apart from the fractional bits, 4 bits are considered for allocating the signed integer part. A drop in accuracy ~ (3-6) % for the SNN classifier depends on time steps/iterations ( $N_{iter}$), as shown in **figure 9(a)**. Hence, a significant amount of computation and power consumption involving the memory is avoided by restricting the number of bits used in the weight's coefficients. **Figure 9(b)** shows the effect on accuracy with total time duration with multiple SNN inference models. It is understood that multiple SNN inference models offer better performance, and such an ensemble of SNN will be an attractive choice for low-power analog SNN inference modules, as discussed in **section 3.4.2**.

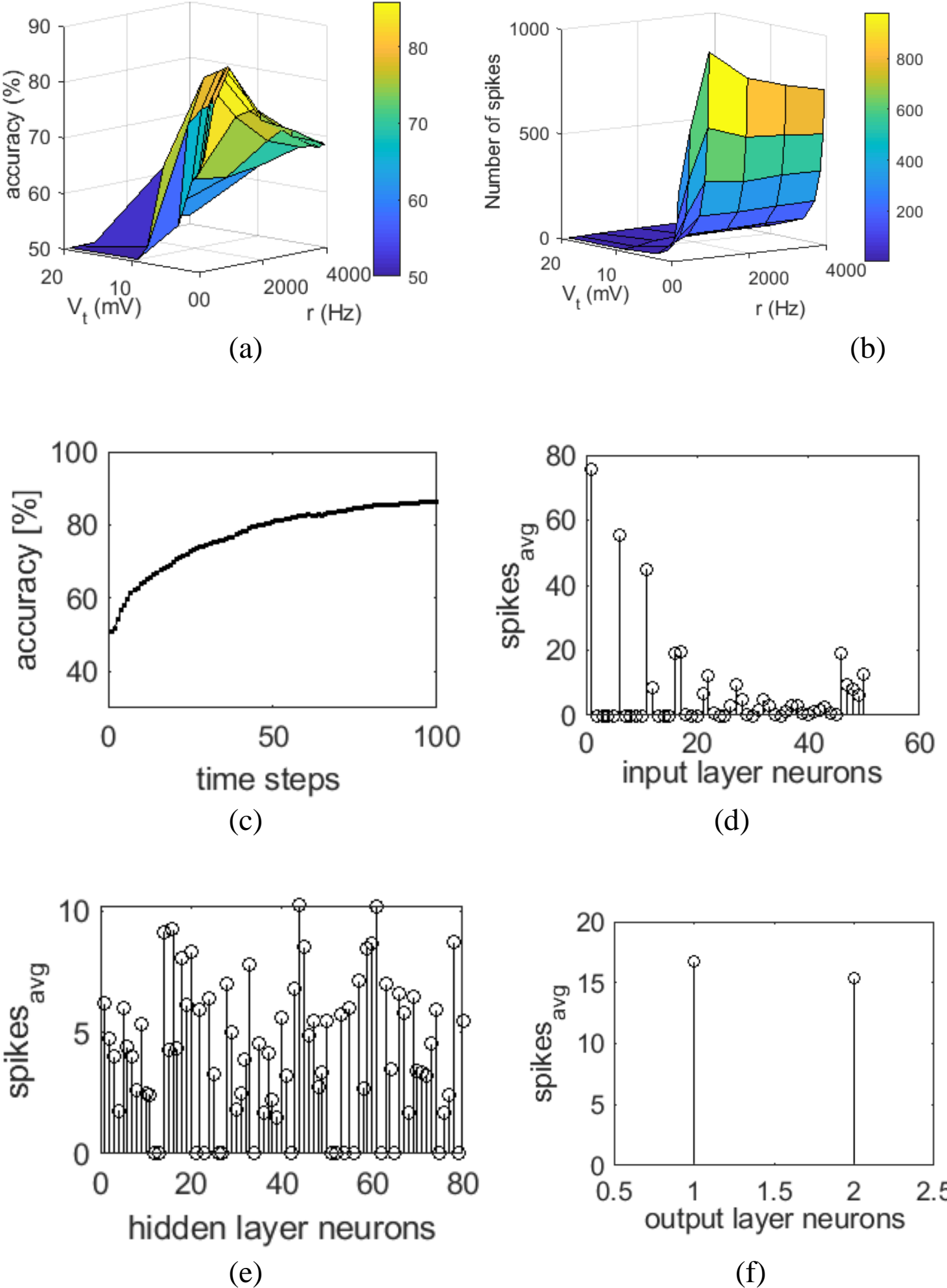

**Figure 8.** (a) Test accuracy w.r.t. time for the SNN classifier (r = 1000Hz and $V_t$ = 1mV) (b) the average number of spikes ($spike_{avg}$) occurring at the hidden layer neurons, (c) hidden layer neurons, (d) output layer neurons, (e) Effect of Vt and r on accuracy, (f) Effect of Vt and r on the number of spikes generated in the hidden layer neuron.

### *3.4 Digital and Analog Implementation of the Proposed Scheme*

#### *3.4.1 Digital Implementation – Architectural schemes for DS FV generation*

The top-level block diagram of the DS generator with SNN inference unit shown in **figure** 10 (a) consists of sub-blocks, viz., i) DS FV generator, ii) slice storage, iii) spike generator, iv) spike address generator, v) weight memory, vi) IF neuron array.

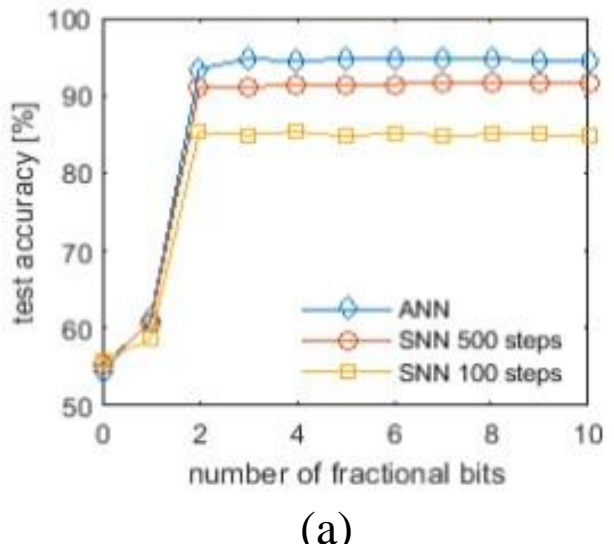

(a)

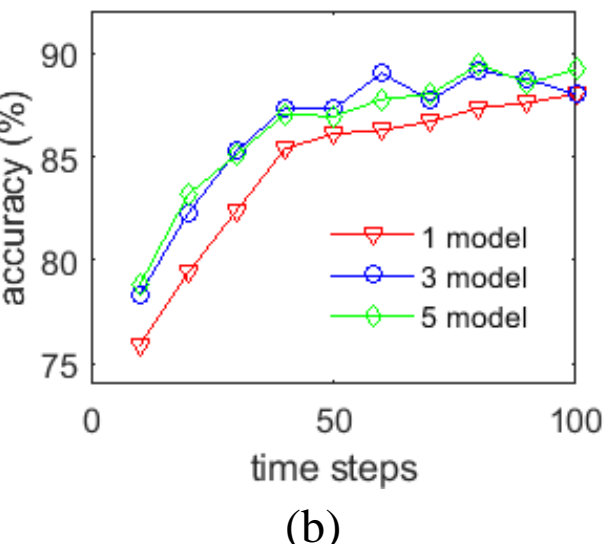

(b)

**Figure 9.** (a) Effect on test accuracy concerning the number of fractional bits in the weight matrix coefficient, (b) Spiking accuracy versus time duration ($T_{dur}$) with 1, 3, and 5 SNN models.

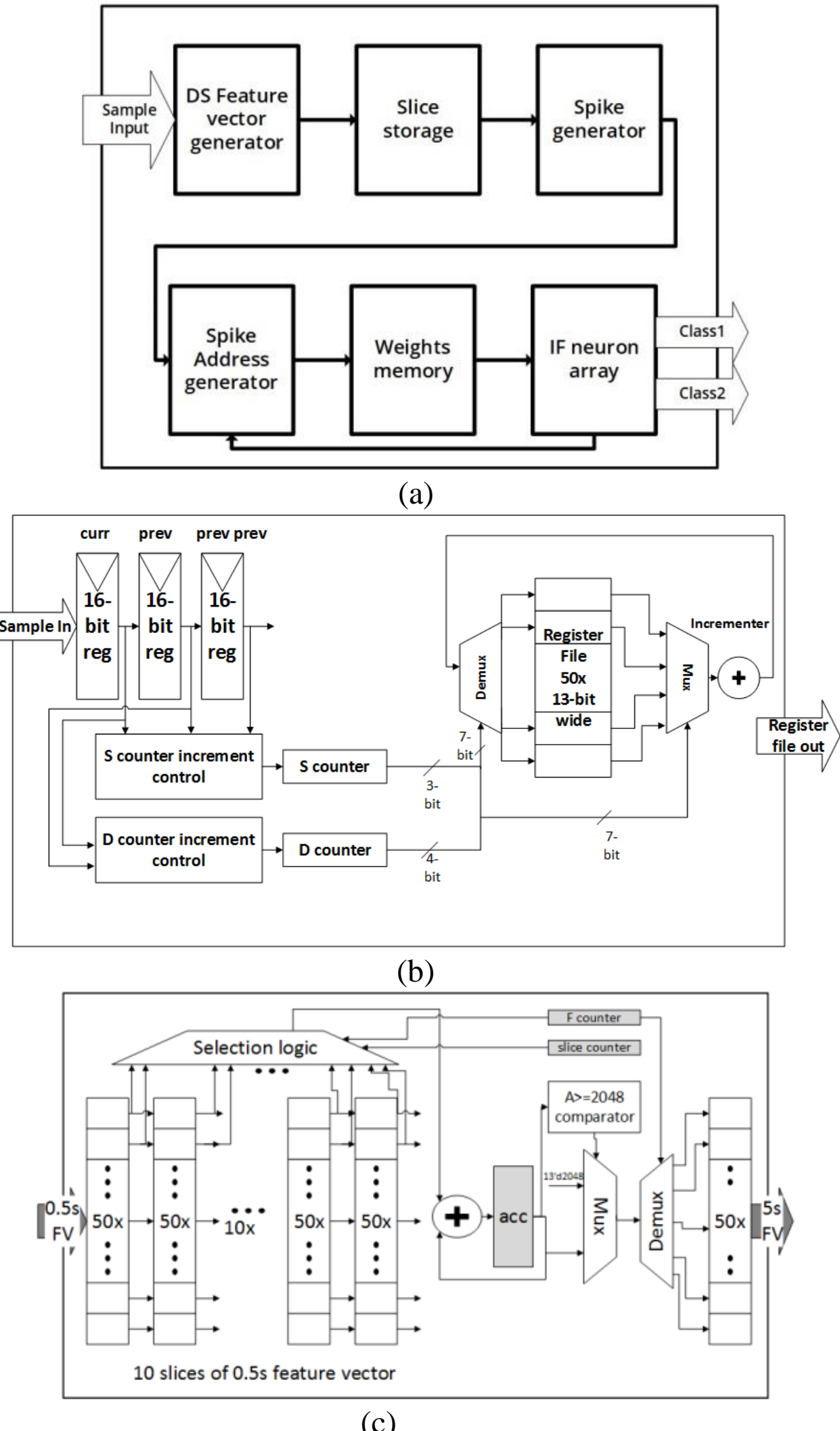

**Figure 10.** (a) Top-level block diagram of the hardware architecture, (b) Feature vector generator, and (c) slice storage scheme.

We implemented the system consisting of DS FV followed by ANN/SNN. **Figures 10 (b) and (c)** show the DS FV generator and the slice storage scheme. The DS generator computes and stores the DS FV every 0.5s in a register. The bit width of the register is chosen to be 8 bits which accommodate up to 256 occurrences of a (D, S) combination pair. The slice storage consists of 10 slices of 0.5s FV unit, which are summed up every 0.5s to generate the input FV of 5s duration fed to the inference unit.

A zero-crossing is detected by comparing the sign bit of the current and previous samples using the XOR gate. For detecting S, three consecutive samples are compared. Counters calculate the number of D and S present in an epoch. The $D_{max}$ and $S_{max}$ values are fixed to 10 and 5 (refer to **section 3.2.4**). Therefore the counter freezes after reaching the maximum D and S values. The counters are reset upon detection of zero crossings (new epoch).

**Figure 11** shows the spike generator, which converts the $FV_i$ to spike patterns according to equation (4). A pseudo-random number (PN) generator is designed using an 11-bit linear feedback shift register (LFSR) for emulating the random number generation, as shown in **figure** 11(b). Each feature vector element value is to be compared with a random number. This is achieved by comparing the outputs of 50 such pseudo-random number generators with FV of dimension 50x1 to get the 50-bit input spike vector in one clock cycle, as shown in **figure** 11 (a). The address location of the weights is accessed based on the spike pattern of the 50-bit FV using a leading zero counter (LZC) [10]. The functioning of an LZC with the highest bit reset logic for an 8-bit spike pattern is shown in **figure 11 (c)**, which produces the output as the location of a spike in the FV, then resets it to scan the following location. The output of the LZC is invalid when all the spikes present in the FV are reset according to the highest bit reset logic. **Fig. 11 (d)** shows that the spike address generator uses the LZC to generate the spike address.

The structure of an I&F and MAC neuron is shown in **figures 12 (a) and (b),** respectively. The weight coefficients between the I&F neurons and the neurons in the previous layer producing spikes are added and stored in an accumulator. Suppose the accumulated value exceeds $V_t$, the IF neuron generates a spike used by the address generator of the next layer. The structure of the IF neuron array for an SNN is shown in **figure 12 (c)**.

### *3.4.2 Mixed Implementation*

The mixed implementation has significant benefits in terms of energy efficiency with proper design parameters such as voltages and currents [50]. The SNN can be implemented in a resistive crossbar network (RCN) fashion using a memristor composed of CMOS or ferroelectric devices to store synaptic weights [24, 25, 52, 53, 54, 56]. Multi-level analog weights have been implemented using a memristor [52, 53, 61]. The SNN model could be mapped directly using the physical characteristics of nano-scale transistors and memristive devices instead of artificial digital emulation, and hence provides significant energy savings compared to the digital implementations [51, 58].

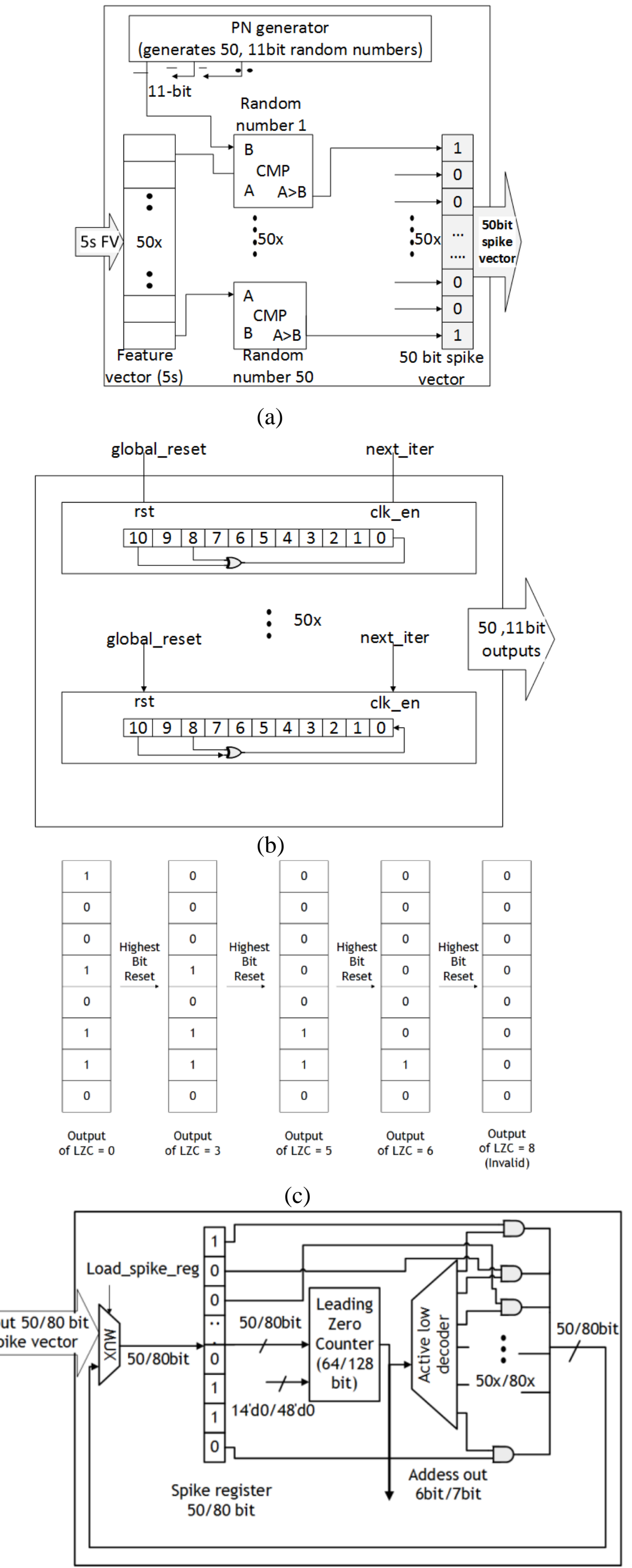

**Figure 11.** (a) Spike generator by comparing 50-bit FV with 50-bit random number generated using LFSR, (b) pseudo-random number generator, (c) Illustration of a leading zero counter (LZC) and highest bit reset logic, (d) Spike address generator with LZC and highest bit reset logic.

3.4.2.1 Mixed implementation of feature extraction

The mixed implementation of the FV followed by a multi-bit RCN is shown in **figure 13(a)**. It consists of a differentiator, comparator, counters, decoder, digital to analog converters (DACs), current-controlled oscillators (CCOs), and multi-bit RCN as the major sub-modules.

Low-power implementation of the front-end amplifier and the CCO has been previously studied [31, 32]. We used a current controlled oscillator (CCO) with three inverter stages for a frequency range of 1 kHz to 650 kHz, as shown in **figure 13(b)**. The difference between the maximum and minimum frequency levels decides the quantization levels in the binary spike frequency. Hence, the precision of the amplitude of the input FV is captured in the form of binary spike frequency. It can be seen from **figure 13(c)** that the aforementioned oscillating frequency is achieved with a low input current (<300 nA) range.

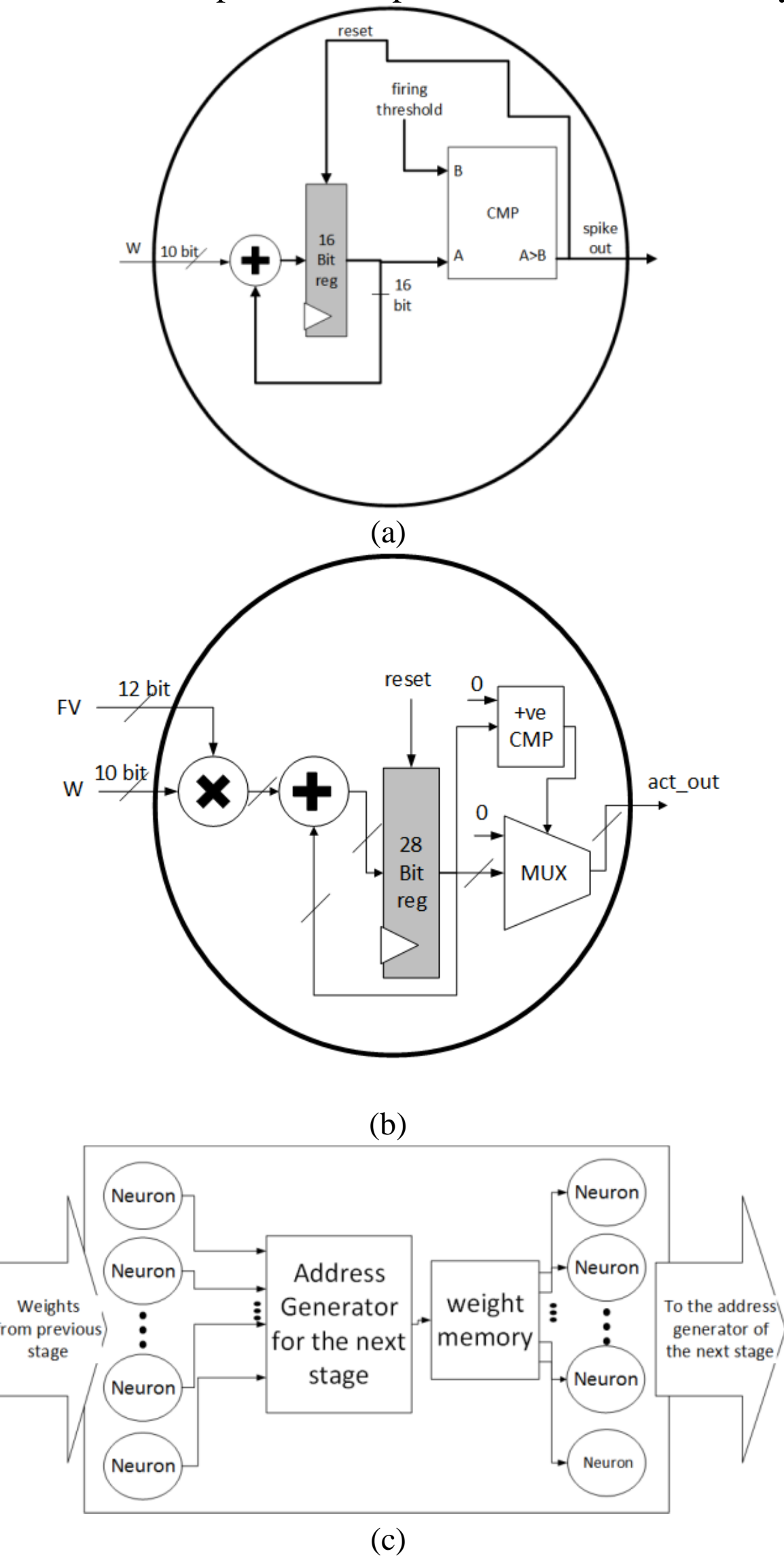

**Figure 12.** (a) IF Neuron (b) MAC neuron (c) IF Neuron array

The RCN network with the memristive weights and the output neuron circuit is shown in **figure 13(d).**

3.4.2.2 Variation analysis for the analog design of the RCN

The neural network implemented in an analog fashion considers the memristors for representing the weights. This variation of the resistive weights is inevitable in analog implementations. Hence, the performance of the NN inference module, including the variations, must be assessed. We studied the trained weights' variation analysis (6 bits taken, excluding sign bit) to determine analog implementation feasibility. It is to be noted that the range of trained weights lies between 10.25 and 0. The histogram of the random

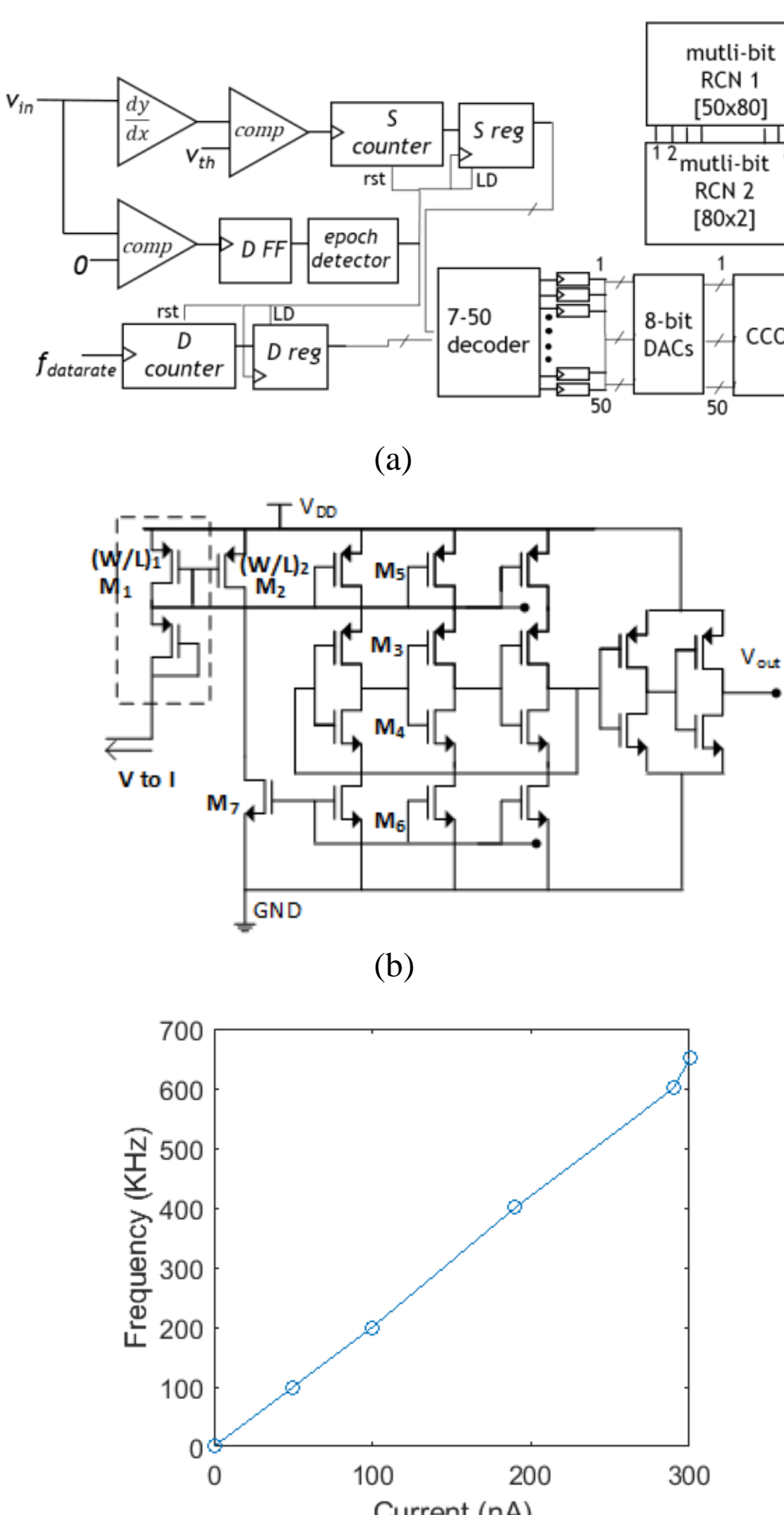

(a)

(b)

(c)

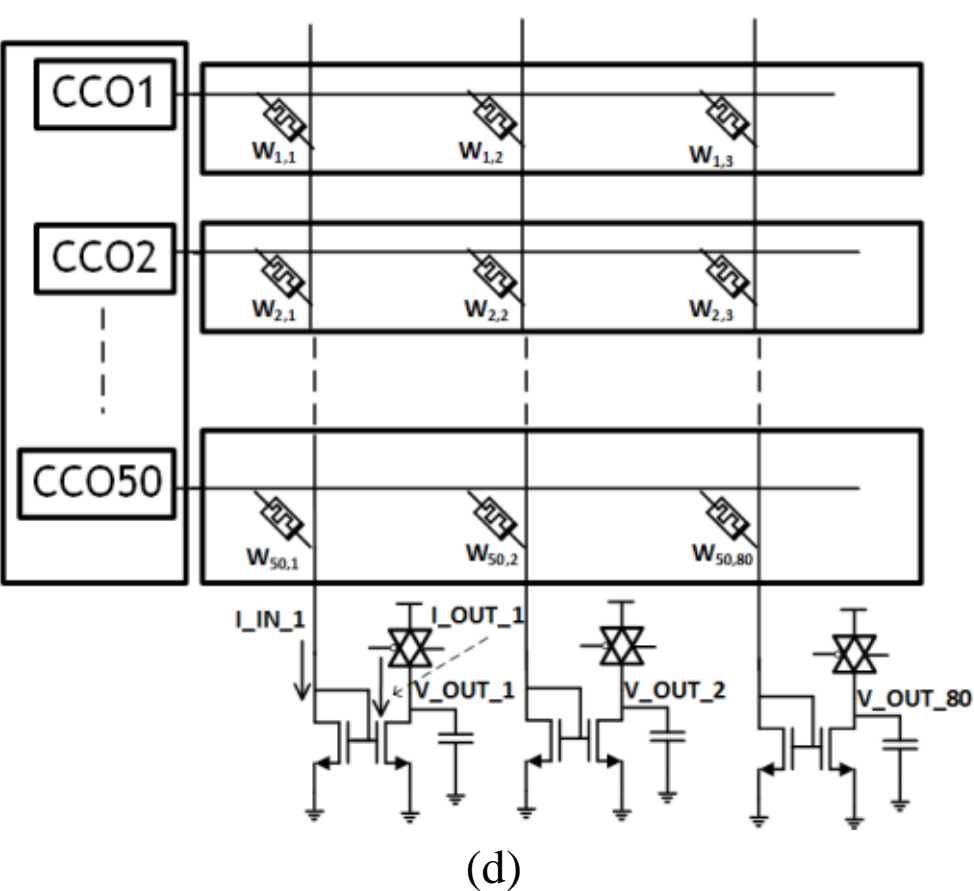

(d)

**Figure 13.** (a) Block-level diagram for the analog scheme for DS FV generation followed by SNN inference, (b) Circuit diagram of a current controlled oscillator (CCO), (c) The input current versus the oscillating frequency characteristics of the CCO, (d) Illustration of resistive crossbar network (RCN) with the output neuron circuit.

Gaussian variation in the trained weights is shown in **figure 14(a)**, which corresponds to a percentage standard deviation of σ=36.68% [52]. We further analyzed the effect on test accuracy versus the percentage standard deviation and found that both the models tolerate up to σ=30%, as shown in **figure 14 (b)**.

3.4.2.3 Crossbar scheme for SNN, using multi-bit memristor weights

The crossbar SNN circuit can be efficiently implemented using multi-bit memristor weights [23, 24, 27, 28, and 29]. A simple crossbar circuit that used programmable weights in the form of the memristor is shown in **figure 14(c)**. From [26], we know that memristor is compatible with CMOS systems. Hence it will be a viable option for compacting the SNN classifier using resistive crossbar memory (RCM). The effect of variation with minimal change in the output neuron threshold potential ($V_t$), as shown in **figure 14(d)**, justifies the scheme's robustness. The gaussian variation ~30% in the output neurons ($V_t$), as demonstrated in **figure 14 (e)(f),** was incorporated to model the variation in transistors in the neuron and assess its effect on test accuracy.

The synapse for the crossbar can be represented using a memristor or multi-bit ReRAMs [21, 52]. The conductance varies from (100 to 900) µS, which supports multi-bit (6 bits) weights [52]. It has been seen that the energy dissipation is restricted if the input bit pattern entering the system is allowed for a short period ($T_{on}$=50ns) [30]. The current flowing through the synapses should be chosen optimally. If the synapse current is too low (<100 nA), then a slight variation in the gate voltage at the PMOS synapse will cause a significant fluctuation in its resistance value due to the exponential dependence on the drain to source voltage in the subthreshold region. Also, if it is too high (>100 nA), the power dissipation will increase. Hence, an optimal value of $I_{synapse}$ should be chosen.

The output neurons store the voltage in a capacitor dedicated to each column, as shown in **figure 13(d)**. Initially, before presenting a new input FV, the capacitors are fully charged. Then based on the input and the conductance of the crossbar junctions, the capacitor is discharged. The winner neuron is the capacitor with the lowest voltage. The capacitance value should be chosen following the maximum amount of charge it will release before the occurrence of the spike. After the spike event, the output neuron goes back to its resting potential, i.e., the capacitor is charged back to the maximum value. A capacitor of 2pF at each column is required to get a voltage range of 0.5V, as calculated from equation (5).

$$C \cdot V_r = Q_{spike} \quad (5)$$

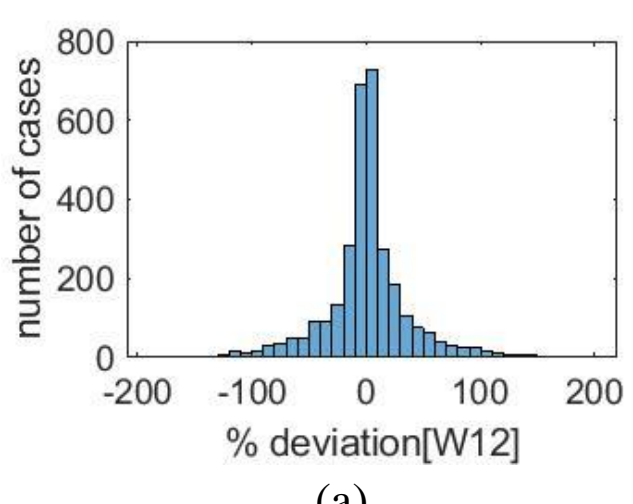

(a)

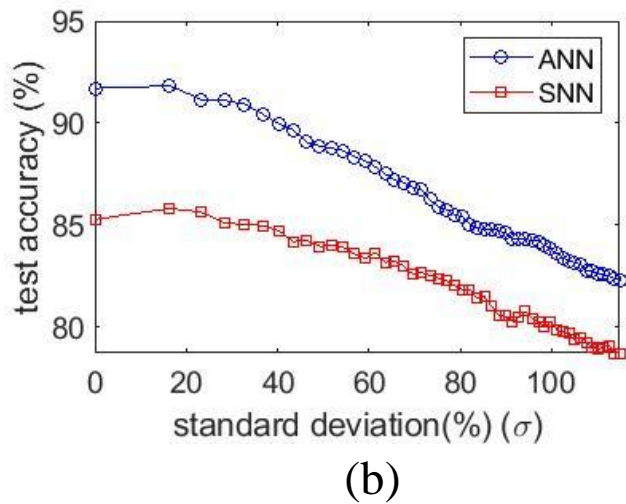

(b)

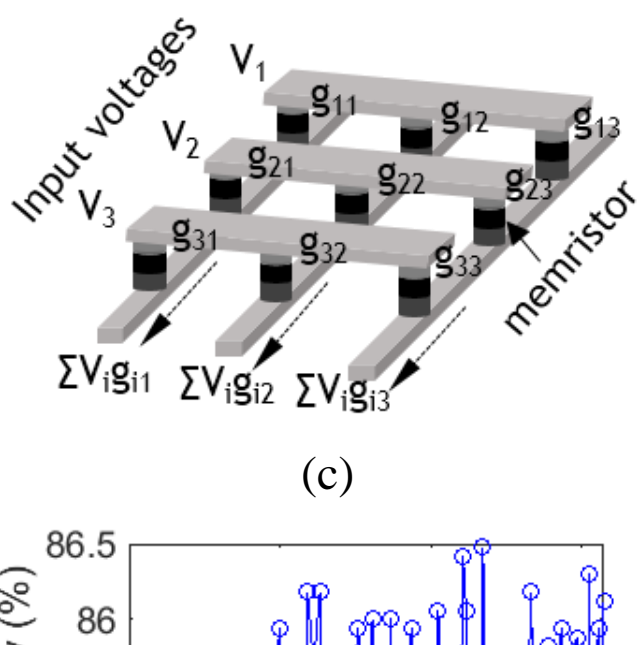

(c)

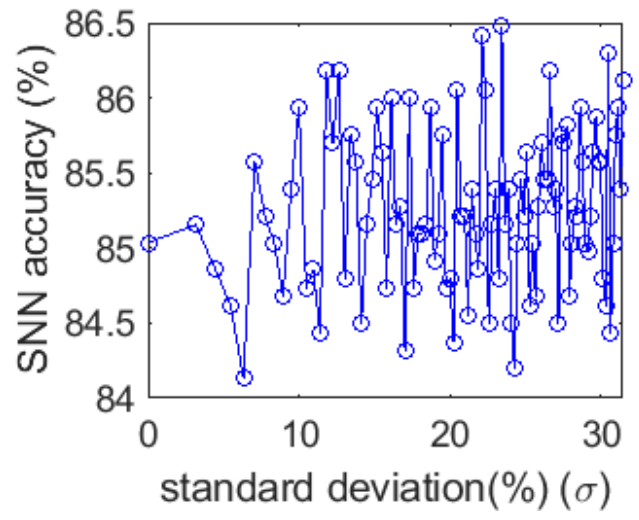

(d)

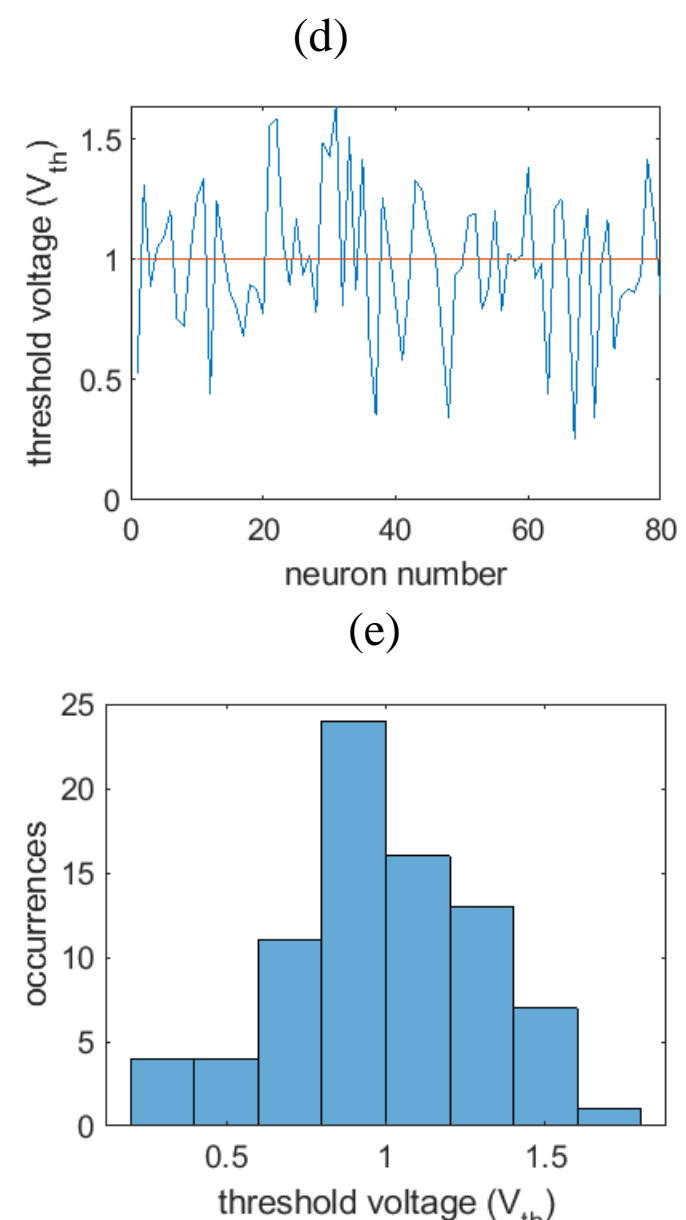

(e)

(f)

**Figure 14.** (a) Histogram of % deviation of weights with standard deviation (σ=36.68%) for incorporating random Gaussian variation, (b) Effect on test accuracy versus % standard deviation of trained weights for ANN and SNN, (c) A representation of resistive crossbar network for computing output current in terms of input voltages and programmable weights, (d) SNN test accuracy versus variation in the threshold potential of the output neuron, (e) The variation in threshold potential of output neurons is shown using boxplot for 30% standard deviation in Vth (f) histogram plot depicting the occurrences are shown using boxplot for 30% standard deviation in Vth.

Here, C, Vr=0.5V, Qspike = 1pC are the capacitance, voltage range across the capacitor, and the amount of charge needed to generate a post-synaptic spike.

## 4. Results and Discussions

This section compares the time domain DS features with optimized FFT features in terms of test accuracy, resources consumed, and power dissipation. Then we compare the DS-SNN system with the DS-ANN system to assess its resource utilization on an FPGA. The ANN unit requires DSP blocks for the multiplication operations involved, which is not needed for the SNN unit. Next, we synthesized the DS-SNN (digital) in an ASIC platform to assess its area and power consumption. To obtain further benefits, we estimate the power consumption of the analog DS-SNN and compare it with the digital DS-SNN.

### *4.1 Conventional methods for audio signal processing*

Conventional methods involve selecting frequency-domain features such as Fast Fourier Transform (FFT) and Discrete Wavelet Transform (DWT) [33-37]. Hence, we compare the performance of the time domain DS FV with the FFT FV scheme regarding classification accuracy. In **figure 15**, we have compared features derived from FFT, DS, and DWT. We have optimized FFT parameters like the number of points, bit precision and computed it by three different approaches, viz., 1) by computing an N=65536 point FFT over the 5s duration and then averaging the bins for feature dimension reduction (Case I), 2) by considering a window size (200ms) with no overlap and computing N=4096 point FFT over the segment at each window and selecting the maximum value of PSD among all windows followed by FV dimension reduction by bin averaging (Case II), 3) by considering window size (50 ms) throughout the 5s segment on which an N point FFT is performed (N=512, 256, 128, 64, 32, 16), only the max value of PSD among all windows is considered in the FV (Case III). The FV derived from the FFT (Case III) is convenient compared to the former two cases because N<512-point FFT is computed, eliminating the FV dimension reduction process.

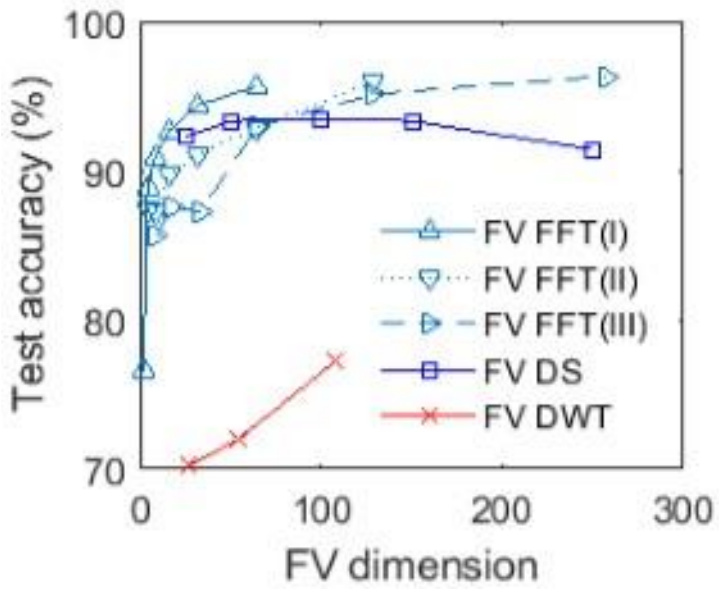

**Figure 15.** Effect on test accuracy for different feature vectors (FV).

Apart from FV derived from FFT, FV is also derived from DS and discrete wavelet transform (DWT), derived from the time domain. The comparison is made between the different FVs while keeping the FV dimension similar. The size of the FV decides the number of nodes at the input layer of the neural network. The computational complexity of FVs derived from the frequency domain will be more than those derived from the time domain.

## 4.2 Comparison between 128-point FFT FV (optimized) and DS FV

The performance of DS FV is more suitable in terms of performance and computational complexity. A comparison of power and resource utilization using DS and 128-pt FFT FV targeting Zynq AP SoC XC7Z020 CLG484-1 FPGA is shown in **Table 2**, from which it is inferred that the FFT FV takes ~3x more power compared to DS FV and consumes more resources. As the FFT-based scheme's power consumption is more than DS FV with similar accuracy, we have chosen the low complexity time-domain DS FV for our application. Static power is identical due to the similar memory size used for storing the FVs. Dynamic power is significantly higher, as FFT involves significantly more arithmetic computations.

## 4.3 Digital schemes for DS-ANN and DS-SNN

System-level architectures of the DS FV generator and ANN/SNN inference unit have been designed while targeting Zedboard with Zynq AP SoC XC7Z020-CLG484-1 FPGA.

The architectural implementation of DS FV, followed by ANN/SNN as explained in **section 3.4.1**, is a synchronous system operating at a clock frequency of 22 kHz. The input data sampling rate is 11 kHz; hence a higher clock frequency of 22 kHz is good enough to complete the feature extraction and classification operations. The FV gets updated every 500ms, which determines the system's throughput.

The format of the trained weight coefficient consists of 10 bits, out of which 6 bits are for fractional, 1 bit for the sign, and 3 bits for integer. The weights are stored in distributed ROM memory. The weights of the first layer are stored in 50 address locations, where each location corresponds to the neuron number in the input layer, i.e., the first location contains all the weight coefficients between the first input neuron with the 80 hidden neurons, resulting in 800 bits per address location. Similarly, the weight memory between the output layer neuron and the hidden layer consists of 80 locations with 20 bits per address location.

The utilization report and power consumption report are mentioned in **Tables 3 and 4**, respectively. The trained parameters are stored in a distributed memory which will infer LUTs and FF. For similar network architecture, the ANN occupies 82 DSP blocks due to multiplication operation, whereas in SNN, DSP blocks are not inferred. It is to be noted that a significant portion of the power is consumed by the Mixed-Mode Clock Manager (MMCM), which converts the inbuilt 100MHz clock to the desired clock frequency of operation.

## 4.4 Synthesis results of DS-SNN (digital)

The estimate of different modules involved in the DS-SNN operated at 22 kHz clock frequency in the 65nm technology node is shown in **Table 5**. The computational block (refer to **Fig. 10 (b)(c)**) updates the DS feature vector every 0.5s and produces an output DS feature vector of 5s, which is the input to the classifiers. It consumes ~6.9 µW. The input spike generator (refer to **figure 11**) consists of a PN generator to convert the input feature vector of size 50 to a corresponding 50-bit spike pattern consumed ~0.78 µW.

The IF neuron consumes 26nW with an area of ~298µm$^2$, which is ~4.58x area efficient and ~5x power efficient than the MAC neuron, followed by the ReLU activation function for the first hidden layer (MAC-ReLU_hl) as shown in **Table 6**. It is to be noted that the MAC neuron for the output layer (MAC_ol) will consume more area and power due to an increase in the size of the input bit width. In contrast, the same IF neuron is used in the output layers as the input bit width depends on the bit size of the trained parameters and the average number of strong connections in the previous layers, which are similar to our network configuration (50-80-2).

The number of weights required for our network is equal to 4160. An SRAM memory for storing the weights will consume ~1.66 µW with an area of ~1915 µm$^2$. The logic for selecting the address corresponding to the stored weights based on the spike patterns (**Figure 11(d)**) for the hidden and output layers consumes 1864 µm$^2$ area and ~150nW power. It is concluded that a significant amount of area and power is saved for the SNN inference unit compared to the ANN inference unit when operated at the same clock frequency.

In **Table 7**, we have the distribution of resources and power of the DS-SNN system. The dominant part of the DS FV is the inter FIFO registers (**Figure 10(c)**). Further, the combinational and non-combinational logic ratios required in the DS-SNN for the area and power distribution are 1.2 and 2.45, respectively, indicating the dominance of sequential logic in the system. The dynamic energy of the DS FV and the inference module amounts to be ~29nJ and ~173nJ, which does its computation in 23ms and 132ms, respectively.

### *4.5 Estimation of power consumption by the mixed scheme*

The breakup of the dynamic power consumed by the different blocks in the mixed scheme (refer to **figure 13**) is given in **Table 8**. These results have been obtained from SPICE simulations. The energy dissipated by the differentiator and comparator is ~3nJ for 500ms. The time to arrive at a new FV is termed the evaluation time ($T_{eval}$), equal to 500ms. The power dissipation of the digital blocks D counter (4 bit), S counter (3 bit), and the 7 to 50 decoders followed by 50 counters (8 bit) are obtained from synthesis results using Synopsys Design Vision Tool at 65 nm technology. The equivalent energy dissipation during the evaluation time is mentioned in **Table** 8.

The energy consumed by 50 simple binary-weighted current DAC was estimated to be 0.04nJ from simulations (refer to equation (6)).

$$P_{DAC} = C \times V_{DD}^2 \times f_{DAC} \times n_{bits} \times n_I \tag{6}$$

Here, $C = 0.1fF$, $V_{DD} = 1V$, $f = 2Hz$, $n_{bits} = 8$, $n_I = 50$ are the bit capacitance, supply voltage, frequency of conversion, number of bits, and the number of input layer neurons, respectively.

The energy consumed by the CCO is expressed as in equation (7) and is verified through SPICE simulations.

$$E_{CCO} = n_{cco} \times I_{cco(avg)} \times V_{DD} \times T_{eval} \tag{7}$$

Here, $n_{cco} = n_I = 50$, $I_{cco(avg)} = 20nA$, $T_{eval} = 0.5ms$, are the number of CCOs, the average current through the CCO, the evaluation time required by the CCO.

The synapses in the SNN crossbar will have variations in conductances. The average number of input spikes generated and the conductance of the synapses will determine the current flowing through each row. The energy dissipated will depend on the time duration ($T_{on}$) for which the input bit spike is presented and is expressed as given in equations (8) and (9) for RCN1 and RCN2, respectively.

$$E_{RCN1} = I_{\max} \times V_{DD} \times T_{on} \times \sum_{r=1}^{n_I} \sum_{c=1}^{n_H} n_{spike(avg)_r} \times \left( \frac{G_{syn_{(r,c)}}}{G_{syn(\max)}} \right) \tag{8}$$

$$E_{RCN2} = I_{\max} \times V_{DD} \times T_{on} \times \sum_{r=1}^{n_H} \sum_{c=1}^{n_O} n_{spike(avg)_r} \times \left( \frac{G_{syn_{(r,c)}}}{G_{syn(\max)}} \right) \tag{9}$$

Here, $n_{spike(avg)_r}$, $n_I = 50$, $n_H = 80$, $n_O = 2$, $I_{\max} = 300nA$, $V_{DD} = 1.2V$, $T_{on} = 50e-9s$, $T_p = 5e-6s$, $G_{syn_{(r,c)}}$, $G_{syn(\max)}$ are the average number of spike occurrences in the $r^{th}$ row, input, hidden, output neurons, maximum synaptic current, supply voltage, ON duration of the spike, pulse duration of a spike, the conductance of the synapse, and maximum synapse conductance in the RCN, respectively. The estimated energy dissipation of the rows in the crossbar circuit based on the statistics of the weight distribution and the average number of input spike occurrences is shown in **figure 16**.

In the idle state (absence of spike), the current source supplying the current will dominate the resistance. The OFF current source will have higher resistance ($R_{CS(off)}$) in the order of 100GΩ, which is in series with the synaptic resistances. Hence, the leakage power given by $N_{rows}$ will be given by equation (10)

$$P_{leakage(RCN)} = \left( V^2_{DD} / R_{CS(off)} \right) \times N_{rows} \tag{10}$$

The leakage energy dissipated for the RCN1 and RCN2 is 0.3pJ and 0.48pJ, respectively. This is added to the computation energy in equations (8) and (9) to get the total energy of 0.051nJ and 0.0056nJ, respectively.

(a)

(b)

**Figure 16.** Energy dissipation by each row of (RCN) during the SNN inference (a) RCN1, (b) RCN2

### *4.6 System-level comparison between digital and mixed scheme (DS-SNN)*

**Table 9** shows the comparison between the digital and mixed schemes for the DS-SNN system. It is seen that the mixed scheme is ~30x energy efficient than its digital counterpart. We also compare the energy consumed by the SNN inference unit with other spike-based processors in the literature, as shown in **Table 10**. Therefore, the energy dissipated per spike is estimated using equation (11)

$$E / spike = \left( \frac{E_{tot}}{num_{spikes}} \right) \tag{11}$$

Here, $num_{spikes} = 700$ is the average number of spike occurrences during the inference process (refer to **Section 3.3, figure 8**). It is to be noted, $num_{spikes}$ depends on the total number of time steps required in the inference process. Hence, applications which require less inference time will be more energy efficient.

Table 2: Comparison of power and resource utilization between D and S and 128-pt FFT Feature Vector (FV) targeting ZYNQ AP SOC XC7Z020 CLG484-1 FPGA

| FV type | Power (mW) | | | Resource utilization (%) | | | |
|---|---|---|---|---|---|---|---|
| | Static | Dynamic | Total | LUT | FF | BRAM | DSP |
| D and S | 108 | 116 | 224 | 0.8 | 0.14 | 3.93 | 0 |
| FFT | 115 | 531 | 646 | 6.38 | 1.88 | 6.07 | 6.36 |

Table 3: Resource utilization for a parallel implementation of DS FV generator followed by ANN/SNN architecture

| Resource | Utilization | | Available | % utilized | |
|---|---|---|---|---|---|
| | ANN | SNN | | ANN | SNN |
| LUT | 6784 | 6572 | 53200 | 12.75 | 12.35 |
| FF | 8061 | 10811 | 106400 | 7.58 | 10.16 |
| DSP | 82 | 0 | 220 | 37.27 | 0 |
| IO | 23 | 23 | 200 | 11.50 | 11.50 |
| BUFG | 3 | 3 | 32 | 9.38 | 9.38 |
| MMCM | 1 | 1 | 4 | 25 | 25 |

Table 4: Power consumption distribution

| Power (mW) | | | |
|---|---|---|---|
| Type | | % consumed | |
| | | ANN | SNN |
| Static | | 125 (44%) | 107 (55%) |
| Dynamic | clocks | <1 (<1%) | <1 (<1%) |
| | Signals | <1 (1%) | <1 (<1%) |
| | Logic | <1 (<1%) | <1 (<1%) |
| | DSP | <1 (<1%) | - |
| | MMCM | 101 (99%) | 101 (99%) |
| | I/O | <1 (<0%) | <1 (0%) |
| Total | | 242 | 208 |

Table 5: Estimation of power/area of the DS-SNN system based on ASIC synthesis results

| Module | Area ($\mu m^2$) | Power ($\mu$W) |
|---|---|---|
| Computational block (DS FV generator) | 89066 | 6.89 |
| IF neuron | 298 | 0.026 |
| PN comparator | 8155 | 0.78 |
| Memory | 1915 | 1.66 |
| Spike address generator | 1864 | 0.15 |
| Control FSM | 1050 | 0.063 |

Table 6: Comparison between MAC-ReLU and IF neuron

| Neuron | Area ($\mu m^2$) | Power (nW) |
|---|---|---|
| MAC-ReLU_hl | 1362 | 105 |
| MAC-ReLU_ol | 2563 | 206 |
| IF | 298 | 26 |

Table 7: Estimate of distribution of resources and power of the DS SNN system

| Module | | Area (µm2) | Power (µW) |
|---|---|---|---|
| DS FV | Computation | 39255 | 1.81 |
| | Storage (register) | 49811 | 5.1 |
| SNN inference | Computation | 15151 | 0.85 |
| | Storage (register +SRAM) | 15073(13158+1915) | 3.12 (1.46+1.66) |
| DS SNN (total) | | 1,19,290 | 10.88 |

Table 8: Energy distribution for the mixed scheme

| Modules | Dynamic Energy ($P_D \times T$) (nJ) |
|---|---|
| Differentiator, Comparator (zero crossing, maxima/minima detector) | 3 |
| D counter (4 bit) | 0.305 |
| S counter (3 bit) | 0.150 |
| 7-50 decoder+50 counters | 0.735 |
| 8-bit DACs (50) | 0.04 |
| CCOs (50) | 2.5($T_{eval}$=2.5ms) |
| multi-bit RCN 1 (50 x 80) | 0.051 |
| multi-bit RCN 2 (80 x 2) | 0.0056 |
| Total | 6.8 |

Table 9: Comparison between digital and analog scheme

| Modules | Energy (dynamic power) (nJ) | |
|---|---|---|
| | Digital | Analog |
| DS FV | 29.44 | 6.73 |
| SNN Inference | 173.32 | 0.057 |
| Total | 202.76 | 6.8 |

Table 10: Estimation of energy consumption during SNN inference based on different spike processors in literature

| Reference | Energy(E)/spike, (pJ/spike) | Energy dissipated during SNN inference ($E \times N_{spk}$), pJ | Application (technology) |
|---|---|---|---|
| 2003([15]) | 2850 | 19,95,000 | Analog model of leaky IF neuron (1.5µm) |
| 2011([17]) | 45 | 31,500 | 256 digital IF neuron (45nm SOI) |
| 2012([18]) | 0.4 | 280 | STDP learning rule circuit (90nm) |
| 2015([19]) | 9.3 | 6510 | CMOS spiking neuron (180nm) |
| 2017([20]) | 0.004 | 2.8 | Morris-Lecar artificial neuron (65nm) |
| 2018([21]) | 0.014-1.4 | 9.8-980 | CMOS memristive synapse for NeuroSoC system (130nm) |
| 2018([22]) | 4.3 | 3010 | CMOS neuron and memristor crossbar array (MCA) (90nm) |
| This work (digital) | 247.6 | 173320 | SNN using digital circuits (65nm) |
| This work (mixed) | 0.081 | 57 | MCA for analog SNN (65nm) |

## 5. Conclusion

This work presented a simulation-based study of human footstep acoustic classification in natural surroundings for wireless sensor nodes, using simple time-domain features using an SNN classifier. We have considered bio-inspired and approximate computing techniques to design the power-constrained sensor node. An appropriate choice of $f_s$ and input feature vector, i.e., (D, S), was determined, such as to achieve an acceptable accuracy (>90%) with fewer computations involved in the feature extraction process. A time-domain-based (D, S) FV showed ~3x power benefits compared to the frequency domain FFT FV. Implementation results of the DS ANN and DS SNN system were carried out targeting Zedboard with Zynq AP SoC XC7Z020-CLG484-1 FPGA to find that the latter requires fewer computational resources (no DSP blocks) compared to the former, which consumed ~38% of the available DSP modules. The proposed spike-based DS-SNN digital ASIC requires ~0.12mm$^2$ area and 10.88μW dynamic power estimated using the Synopsys Design Vision tool. Moreover, the proposed mixed scheme for the DS-SNN system consisting of low-power analog modules for generating the DS FV followed by a multi-level RCN for inference shows ~30x energy benefit compared to the digital scheme. Although SNN provides low-power operation, at the algorithm level itself, ANN to SNN conversion leads to the inevitable loss of classification accuracy, which was found to be around ~5%. To overcome the loss in accuracy, we exploited the low-power operation of the analog processing SNN module and applied redundancy and a majority voting scheme. We have compared our SNN inference scheme with other spike-based processors. It is inferred that spike-based processors are suitable for energy-efficient applications, provided the performance is within the permissible limits.

The acoustic WSN studied here will act as the 1st stage of the security surveillance system, which will activate the 2nd stage (consisting of cameras capturing videos/images) for further assessment of the scene. In future work, low-power feature extraction schemes on the frame of images captured by the camera data followed by a bio-inspired SNN classifier could be undertaken.